\pdfoutput=1
\documentclass[11pt]{article}

\usepackage[letterpaper,margin=1in]{geometry}
\usepackage[T1]{fontenc}
\usepackage[utf8]{inputenc}
\usepackage{tgtermes}

\usepackage{amsmath,amsfonts,bm}

\def\eqref#1{equation~\ref{#1}}

\def\1{\bm{1}}

\DeclareMathAlphabet{\mathsfit}{\encodingdefault}{\sfdefault}{m}{sl}
\SetMathAlphabet{\mathsfit}{bold}{\encodingdefault}{\sfdefault}{bx}{n}

\usepackage{amssymb}
\usepackage{booktabs}
\usepackage{nicefrac}
\usepackage{microtype}
\usepackage{xcolor}
\usepackage{graphicx}
\usepackage{placeins}
\usepackage{multirow}
\usepackage[font=small,labelfont=bf]{caption}
\usepackage{subcaption}
\usepackage[round]{natbib}
\usepackage{authblk}
\usepackage[colorlinks=true,linkcolor=blue!60!black,citecolor=blue!60!black,urlcolor=blue!60!black]{hyperref}
\usepackage{url}

\renewcommand{\topfraction}{0.95}
\renewcommand{\textfraction}{0.05}
\renewcommand{\floatpagefraction}{0.8}

\newcommand{\LFHF}{LF$\rightarrow$HF}

\title{Multi-Fidelity Training Reshapes Uncertainty Decomposition in Atomistic Models}

\author[1]{Matteo Tolloso}
\author[1]{Davide Bacciu}
\author[2]{Luigi Bonati}
\affil[1]{Department of Computer Science, University of Pisa}
\affil[2]{Atomistic Simulations, Italian Institute of Technology}
\affil[ ]{\small\texttt{matteo.tolloso@phd.unipi.it}, \texttt{davide.bacciu@unipi.it}, \texttt{luigi.bonati@iit.it}}
\date{}

\begin{document}

\maketitle

\begin{abstract}
Multi-fidelity training, which combines abundant low-fidelity data with scarce high-fidelity labels, is increasingly common in atomistic modeling, but its effect on predictive uncertainty remains largely unexplored. We investigate this question using deep ensembles of mean--variance MACE models trained on ANI-1x, contrasting high-fidelity training from scratch with low-fidelity pretraining followed by fine-tuning. We characterize aleatoric, epistemic, and total uncertainty throughout training and under two distribution shifts: unseen molecular systems and higher-energy configurations. We find that pretraining reduces prediction error and improves uncertainty-based ranking across all test sets, but does not yield well-calibrated predictive variances; calibration deteriorates more severely under energy extrapolation. Training-dynamics analysis localizes most of this calibration degradation to high-fidelity optimization, during which the learned aleatoric variance becomes overconfident on held-out data. Moreover, multi-fidelity fine-tuning produces a more effective uncertainty-based acquisition function than high-fidelity training from scratch. These findings show that uncertainty ranking and calibration capture distinct properties and must be assessed separately in multi-fidelity atomistic models, especially under extrapolation.
\end{abstract}

\section{Introduction}
\label{sec:intro}

Reliable uncertainty estimation is essential in atomistic machine learning, where models are routinely deployed beyond their training distribution, for instance in active learning and molecular dynamics~\citep{podryabinkin2017active,unke2021machine,perego2024data}. Recent work has investigated ensembles, calibration and the decomposition of predictive uncertainty into aleatoric and epistemic components for atomistic models~\citep{valdenegro2022deeper,tan2023single,busk2023graph,dai2025uncertainty,grasselli2025uncertainty}. Meanwhile, multi-fidelity~\citep{musil2019fast,tran2020multi,kim2024data,gardner2025understanding,ko2025data} and transfer learning~\citep{falk2023transfer,rocken2025enhancing,novelli2025fast,radova2025fine} have emerged as practical strategies for combining abundant low-fidelity (LF) data with scarce high-fidelity (HF) labels, typically obtained at different levels of electronic-structure theory~\citep{smith2019approaching,chen2023data,zaverkin2023transfer}. Yet, how multi-fidelity training shapes predictive uncertainty remains largely unexplored.

In standard statistical settings, the aleatoric--epistemic decomposition separates irreducible data noise from uncertainty about the model~\citep{bishop2006pattern,kendall2017uncertainties}. Its interpretation is less straightforward in atomistic modeling, where reference energies are deterministic and data coverage is highly structured. Moreover, uncertainty is commonly evaluated only at convergence and through a single score, conflating two distinct properties: whether uncertainty correctly \emph{ranks} configurations by prediction error, as required for acquisition and filtering, and whether its magnitude matches the error, i.e.\ whether the error bars are \emph{calibrated}~\citep{kuleshov2018accurate,rasmussen2023uncertain}. These properties need not evolve together across fidelity changes, distribution shifts, or training stages. The latter is particularly relevant in practice because deployment ultimately relies on a checkpoint selected during training.

\textbf{Contributions.}
Using ANI-1x data~\citep{smith2020ani}, with DFT as low fidelity and CCSD(T)/CBS as high fidelity, we compare LF-only, HF-only and pretrain-then-fine-tune (\LFHF) training of mean--variance MACE ensembles. We evaluate two distribution shifts: unseen molecular systems and higher-energy configurations of known systems. Across five independent data splits, we follow aleatoric (AU), epistemic (EU) and total uncertainty (TU) throughout training. Our study uses theoretically grounded mean--variance deep ensembles~\citep{lakshminarayanan2017simple,kendall2017uncertainties} and, to our knowledge, provides the first systematic analysis of how multi-fidelity training affects uncertainty dynamics. We find that:
(i) LF pretraining consistently improves predictive accuracy and the error \emph{ranking} induced by AU, EU and TU across all test sets, but does not always improve \emph{calibration};
(ii) unlike system shift, energy extrapolation decouples ranking from calibration: models rank high-energy errors more effectively while underestimating their magnitude more severely, and in-domain isotonic recalibration does not repair this;
(iii) along training, most of the calibration is lost in the HF stage in both training protocols, and the loss emerges faster after pretraining: AU becomes markedly overconfident on unseen data, whereas EU mitigates the resulting degradation in TU;
(iv) pretraining makes uncertainty a better guide for data acquisition: despite starting from higher accuracy, the pretrained model benefits more from uncertainty-guided selection than a model trained from scratch.
We release splits, protocols, and evaluation code at \url{https://github.com/matteotolloso/mace}. Related work is reviewed in Appendix~\ref{sec:related_work}.

\section{Methodology}
\label{sec:methods}

\begin{figure}[t]
    \centering
    \includegraphics[width=\linewidth]{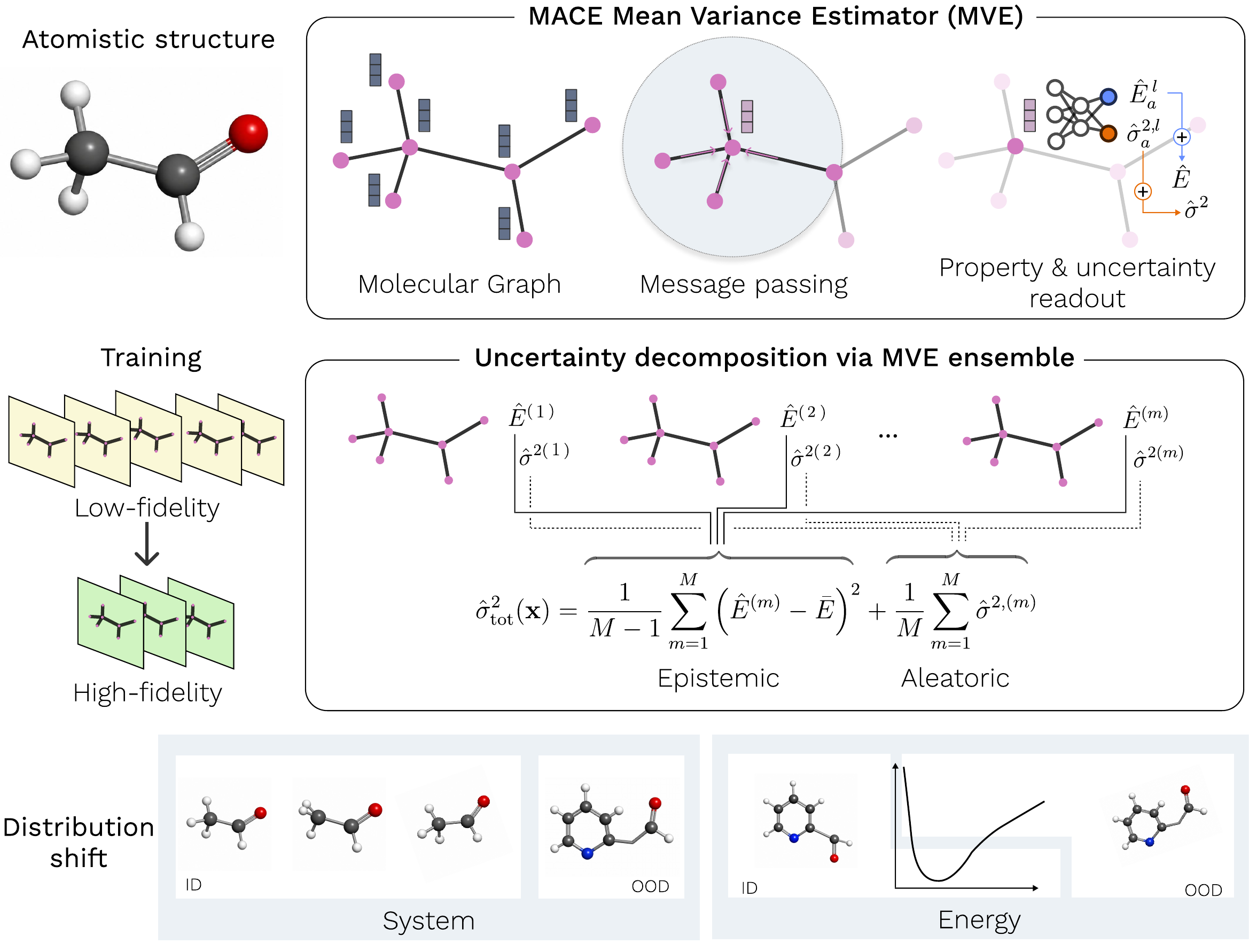}
    \caption{\textbf{Overview of the uncertainty estimation pipeline.} A mean--variance MACE model maps a structure to an energy and a variance. In an ensemble of such models, the average predicted variance gives the aleatoric uncertainty and the disagreement between members the epistemic uncertainty.}
    \label{fig:workflow}
\end{figure}

Figure~\ref{fig:workflow} summarizes our pipeline: an ensemble of mean--variance MACE models, whose average predicted variance and disagreement define the aleatoric and epistemic components, respectively.

\subsection{Error and predictive uncertainty}
\label{sec:error_and_predictive}

For a structure $\mathbf{x}$, defined by its atomic species and positions, with reference energy $y$, predictive uncertainty quantifies the reliability of a prediction $\hat{y}(\mathbf{x})$ before computing $y$. It is represented by the predictive distribution $p(y|\mathbf{x}, \mathcal{D})$ conditioned on training data $\mathcal{D}$~\citep{murphy2012machine}, and it is commonly decomposed into aleatoric and epistemic components~\citep{gawlikowski2023survey}. Epistemic uncertainty captures uncertainty in the predictor, arising from limited data coverage, model misspecification, or extrapolation, and is in principle reducible with additional data. Aleatoric uncertainty is conventionally associated with irreducible observation noise. However, since reference energies of a given electronic-structure method are deterministic~\citep{helgaker2013molecular}, we instead interpret it operationally as input-dependent residual variability, or configuration-dependent difficulty. Whether the two terms actually behave as irreducible and reducible uncertainty is tested in Section~\ref{sec:results:decomposition}.

With model parameters $\theta$ and target $Y$, the law of total variance~\citep{bishop2006pattern} gives
\begin{equation}
\label{eq:total_variance}
\operatorname{Var}(Y | \mathbf{x},\mathcal{D})
=
\underbrace{
\mathbb{E}_{p(\theta | \mathcal{D})}
\!\left[
\operatorname{Var}(Y | \mathbf{x},\theta)
\right]
}_{\text{aleatoric}}
+
\underbrace{
\operatorname{Var}_{p(\theta | \mathcal{D})}
\!\left[
\mathbb{E}(Y | \mathbf{x},\theta)
\right]
}_{\text{epistemic}}.
\end{equation}
With deterministic labels, however, the learned variance captures any residual variation the model fails to explain, whether due to limited data, misspecification or optimization, while model disagreement provides only a proxy for epistemic uncertainty~\citep{hullermeier2021aleatoric}. We therefore use \emph{aleatoric} and \emph{epistemic} as operational labels for the two variance terms rather than as uniquely identified sources of uncertainty (Appendix~\ref{sec:app:theory}).

\subsection{Predictive model and uncertainty parameterization}
\label{sec:predictive_model}

\textbf{Mean--variance MACE model.} We build on MACE, an E(3)-equivariant message-passing architecture with higher-body atomic-cluster-expansion features~\citep{batatia2022mace,batatia2025design}.
Given an atomistic configuration $\mathbf{x}$ with atom set $\mathcal{A}$, MACE represents the total energy as a sum of atomic contributions $a$ across interaction layers $\ell$:
\(
\hat{E}(\mathbf{x})
=
\sum_{a \in \mathcal{A}}
\sum_{\ell=1}^{L}
\hat{E}_a^{\ell}.
\)
To obtain heteroscedastic (input-dependent) uncertainty estimates, we extend the final readout layer to output, for each atom $a$, both an energy contribution and a variance contribution,
\(
(
\hat{E}_a^{L},
\hat{\sigma}_{a}^{2,L}
),
\)
and obtain the structure-level predictive variance as their sum,
\(
\hat{\sigma}^2(\mathbf{x})
=
\sum_{a \in \mathcal{A}}
\hat{\sigma}_{a}^{2,L}.
\)
Modeling the reference energy as
\(
E | \mathbf{x}
\sim
\mathcal{N}
\big(
\hat{E}(\mathbf{x}),
\hat{\sigma}^2(\mathbf{x})
\big),
\)
we train the model by minimizing the Gaussian negative log-likelihood (NLL)
\begin{equation}
\mathcal{L}_{\mathrm{NLL}}
=
\frac{1}{N}\sum_{i=1}^{N}
\Big[
\frac{(E_i - \hat{E}(\mathbf{x}_i))^2}{2\hat{\sigma}^2(\mathbf{x}_i)}
+
\frac{1}{2}\log \hat{\sigma}^2(\mathbf{x}_i)
\Big]
+\text{const}.
\end{equation}
The learned variance is the per-model estimate of aleatoric uncertainty in the ensemble decomposition below.

\textbf{Ensemble-based uncertainty estimation.}
To capture epistemic uncertainty, we employ a deep ensemble~\citep{lakshminarayanan2017simple} of $M$ independently initialized mean--variance MACE models (Figure~\ref{fig:workflow}), closely related to the committee models used for active learning of interatomic potentials~\citep{zhang2019active,schran2020committee,yang2022using}. Each member $m$ produces an energy $\hat{E}^{(m)}(\mathbf{x})$ and a variance $\hat{\sigma}^{2,(m)}(\mathbf{x})$, and the ensemble mean is
\(
\bar{E}(\mathbf{x})
=
\frac{1}{M}
\sum_{m=1}^{M}
\hat{E}^{(m)}(\mathbf{x}).
\)
Treating the members as approximate samples from a posterior over parameters, we approximate the two terms of Equation~\ref{eq:total_variance} with averages over the $M$ models, which gives the total uncertainty (TU)
\begin{equation}
\label{eq:tu}
\hat{\sigma}_{\mathrm{tot}}^2(\mathbf{x})
=
\underbrace{
\frac{1}{M}
\sum_{m=1}^{M}
\hat{\sigma}^{2,(m)}(\mathbf{x})
}_{\text{aleatoric (AU)}}
+
\underbrace{
\frac{1}{M}
\sum_{m=1}^{M}
\left(
\hat{E}^{(m)}(\mathbf{x})
-
\bar{E}(\mathbf{x})
\right)^2
}_{\text{epistemic (EU)}}.
\end{equation}
We evaluate the ranking of AU, EU and TU separately, as three competing signals, while for calibration we focus on TU since it is the predictive variance (Section~\ref{sec:experiments}).

\section{Experimental setup}
\label{sec:experiments}

\textbf{Data and fidelities.} All experiments use the ANI-1x data set~\citep{smith2020ani}, which provides molecular configurations labelled at several levels of theory. The low-fidelity (LF) target is the $\omega$B97x/def2-TZVPP DFT energy and the high-fidelity (HF) target is the CCSD(T)/CBS coupled-cluster energy, which is available for far fewer configurations (the ANI-1ccx subset~\citep{smith2019approaching}). Only energies are used. Each split has 50{,}000 DFT and 5{,}000 CC training configurations, 10{,}000 DFT and 1{,}000 CC validation configurations, and 5{,}000 CC configurations per test set; CC configurations are a subset of the DFT ones.

\textbf{Distribution shifts.} The {\textit{system-level shift}} holds out 40\% of the molecular systems entirely: System-ID tests on unseen configurations of the training systems, System-OOD on the held-out systems (Appendix~\ref{sec:system_split}). The {\textit{energy-based shift}} ranks the configurations of every system by their relative DFT energy, with within-system quantile $q_s(x)\in[0,1]$. Configurations with $q_s \le 0.5$ are used for training, validation and the Energy-ID test, configurations with $q_s \ge 0.55$ form the Energy-OOD test, and the band in between is discarded (Appendix~\ref{sec:energy_split}). 

\textbf{Training protocols.} \textbf{LF-only} models are trained on DFT energies and \textbf{HF-only} models on CC energies, both from scratch for 300 epochs. \textbf{\LFHF} models are LF-only models fine-tuned on CC energies for 100 epochs with a tenfold smaller learning rate; every fine-tuned member starts from the LF-only member with the same split and seed. All protocols use the Gaussian NLL, the same architecture and $M=10$ ensemble members (Appendix~\ref{sec:hyperparameters}). The checkpoint of each member is selected by its validation NLL; test sets are never used for selection. This selection criterion retains HF-only members after 30--111 of 300 epochs and \LFHF\ members after one to three fine-tuning epochs (Appendix~\ref{sec:hyperparameters}).

\textbf{Replication and statistics.} The pipeline, from data splitting to ensemble training, is repeated on five independent dataset splits, the unit of replication. Metrics are computed within a split from its ten-member ensemble and summarized by the five-split mean with a two-sided 95\% Student-t interval (4 degrees of freedom); ENCE, RMSE and ratios use geometric means with log-space intervals.

\textbf{Metrics.} Both errors and uncertainties are normalized per atom, i.e.\ energies are divided by the number of atoms and variances by the square of the number of atoms. {\textit{Accuracy}} is the RMSE of the ensemble mean. {\textit{Ranking}} is measured by the area under the sparsification error curve (AUSE)~\citep{lind2024uncertainty}, which compares discarding configurations by uncertainty with discarding them by their true error and rewards identifying the largest errors (lower is better); {\textit{calibration}} by the expected normalized calibration error (ENCE)~\citep{levi2022evaluating}, $\frac{1}{B}\sum_b |\mathrm{MV}_b-\mathrm{MSE}_b|/\mathrm{MV}_b$ over $B=15$ equal-count bins of predicted variance, with mean variance $\mathrm{MV}_b$ and mean squared error $\mathrm{MSE}_b$ (Appendix~\ref{sec:uq_metrics_appendix}). A bin where the variance overestimates the error contributes less than one, so ENCE above one can only come from overconfident bins. Note that, because TU is the predictive variance and AU and EU are its two components, the three cannot all match the magnitude of the error at once: calibration is expected of TU, while the ENCE of AU and EU only shows how much of the error each part accounts for. Our figures therefore highlight TU when showing calibration. All metrics exclude the handful of Energy-OOD configurations (13--16 of 5{,}000 per split) that are compressed below any interatomic distance seen in training (Appendix~\ref{sec:app:filter}).

\textbf{Research questions.} \textbf{RQ1:} how do AU, EU and TU rank and calibrate at convergence across fidelity regimes? \textbf{RQ2:} do system-level and energy-based shifts affect ranking and calibration differently? \textbf{RQ3:} how do the signals evolve during training, especially across the LF-to-HF transition? \textbf{RQ4:} how does pretraining affect uncertainty-guided data acquisition?

\begin{figure}[ht]
  \centering
  \includegraphics[width=\linewidth]{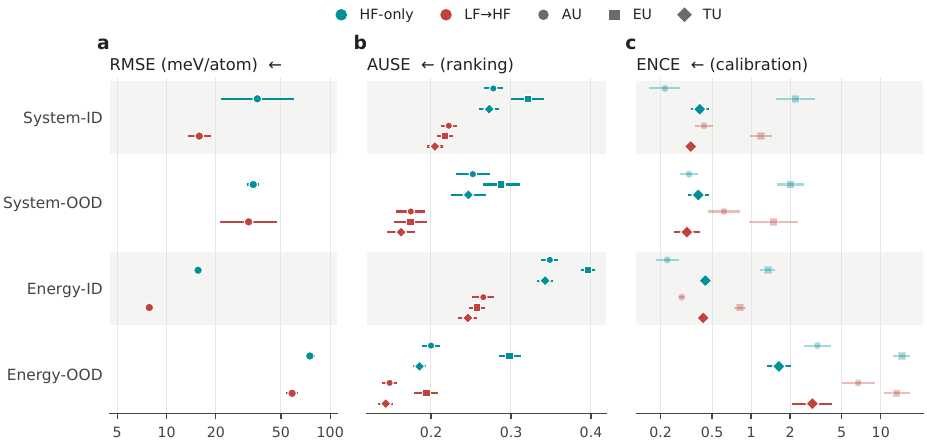}
  \caption{\textbf{Accuracy and uncertainty for the converged models}. \textbf{(a)} RMSE of the ensemble mean, \textbf{(b)} AUSE and \textbf{(c)} ENCE. Colour denotes the protocol (teal HF-only, red \LFHF) and marker the signal (circle AU, square EU, diamond TU); in (c) TU, the predictive variance, is highlighted and its components AU and EU are faded. Bars are 95\% intervals over five splits, and RMSE and ENCE use log axes with geometric means. The wide HF-only System-ID RMSE interval comes from a single configuration in one split, which TU ranks as the most uncertain of 5{,}000.}
  \label{fig:final}
\end{figure}

\section{Results}
\label{sec:results}

We first analyze the converged models (Section~\ref{sec:results:final}). We then examine the full multi-fidelity training trajectory (Section~\ref{sec:results:decomposition}) and compare its high-fidelity stage with HF-only training (Section~\ref{sec:results:hf}). Finally, we present a small acquisition experiment (Section~\ref{sec:results:al}). All results are replicated over five data splits; figures show 95\% intervals, and the complete set of metrics, reliability diagrams and training curves is in Appendices~\ref{sec:supp} and~\ref{sec:app:dynamics}.

\subsection{Pretraining improves ranking but not always calibration}
\label{sec:results:final}

Figure~\ref{fig:final} compares the HF-only and multi-fidelity training. As expected, pretraining reduces the prediction error (RMSE), most strongly in domain but also out of domain. Instead, the other metrics reveal a different behavior of the uncertainty components.  On the one hand, pretraining consistently improves \emph{ranking} (AUSE) (panel b): across all three signals and on all four test sets, pretrained ensembles order prediction errors more effectively than ensembles trained only on HF data. 
Notably, multi-fidelity training reshapes the uncertainty decomposition rather than merely improving prediction accuracy. Indeed, there is a change in the ranking capabilities of the individual components: in HF-only, EU is the weakest ranking signal on every test set; with pretraining, EU ranks as well as or better than AU on three of the four test sets and approaches the ranking quality of total predictive uncertainty (see also Appendix~\ref{sec:supp:spearman}).

By contrast, pretraining has a mixed effect on \emph{calibration} (ENCE; panel c). The relevant uncertainty (TU) is slightly better calibrated after pretraining in domain and on unseen systems, but substantially worse on Energy-OOD. Its components also diverge: EU improves across all test sets, while AU deteriorates. Thus, pretraining lowers error and improves ranking, but does not necessarily produce uncertainties with the correct magnitude.

\begin{figure}[t]
  \centering
  \includegraphics[width=0.7\linewidth]{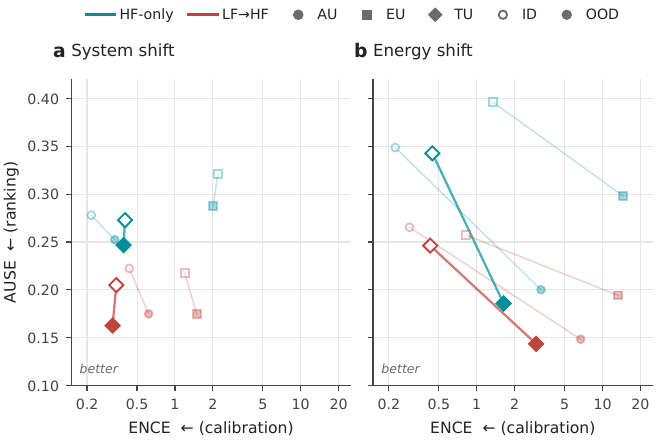}
  \caption{\textbf{Ranking and calibration behavior under different distribution shifts}. ENCE (log $x$) against AUSE ($y$) for the converged models under \textbf{(a)} system shift and \textbf{(b)} energy shift; lower left is best. Colour: protocol; marker: signal, with TU highlighted and AU and EU faded. For every signal a thin line joins the ID (hollow) to the OOD test set (filled).}
  \label{fig:shift}
\end{figure}

Notably, we find that the relationship between ranking and calibration depends critically on the type of distribution shift. To highlight this,  Figure~\ref{fig:shift} places each test set in the calibration--ranking plane and connects every in-domain result to its out-of-domain counterpart. Under system shift, moving from ID to unseen molecular systems improves ranking with little change in calibration. Under energy shift, however, where OOD means higher-energy configurations, ranking \emph{improves} while calibration error increases several-fold: for TU, by approximately $4\times$ with HF-only training and $7\times$ with \LFHF. The Energy-OOD behavior in Figure~\ref{fig:final}c is therefore not specific to pretraining. Both protocols rank high-energy errors better than in-domain errors, yet underestimate them more severely. In other words, the models largely identify \emph{which} high-energy configurations will incur larger errors, since those errors are larger and easier to order, but not \emph{by how much}. Ranking and calibration can therefore move in opposite directions along some extrapolation axes, and isotonic recalibration fitted in domain does not repair this (Appendix~\ref{sec:supp:recal}).

\subsection{Uncertainty signals evolve differently during training}
\label{sec:results:decomposition}

\begin{figure}[t!]
  \centering
  \includegraphics[width=0.98\linewidth]{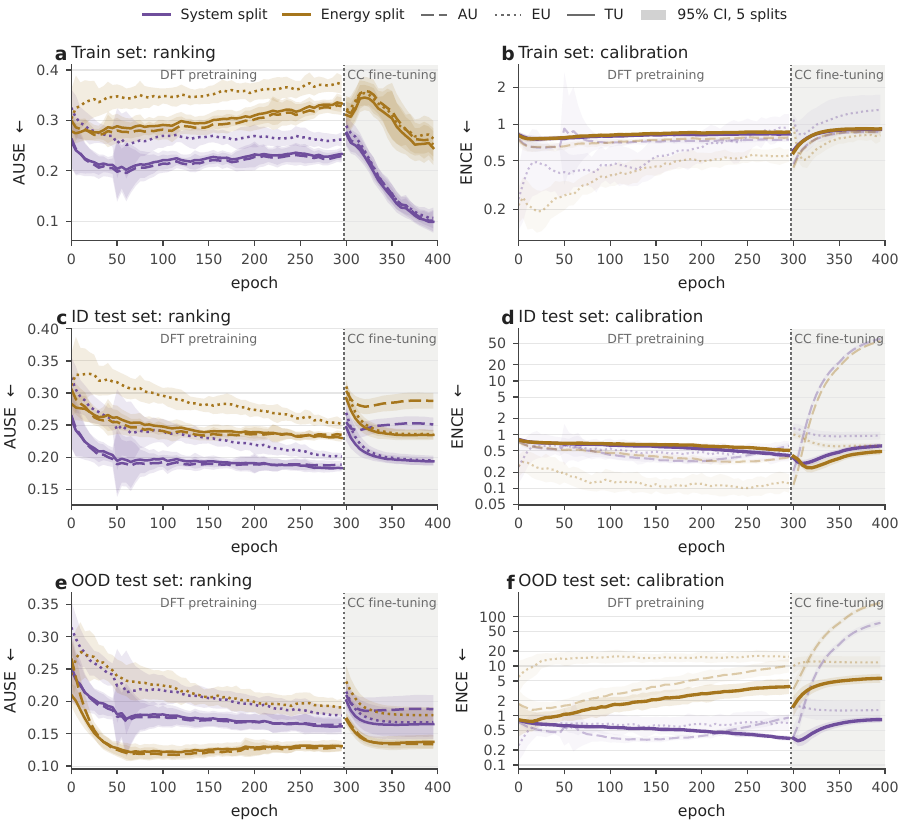}
  \caption{\textbf{Uncertainty evolution during training for the  multi-fidelity case}. \LFHF\ through DFT pretraining (epochs 0--299, scored against DFT labels) and CC fine-tuning (300--399, shaded, CC labels) on its own training data (a, b), the ID (c, d) and the OOD test set (e, f). Note that at epoch 300 the target labels change and, except for Energy-OOD, the evaluated configurations also change; values across the transition are therefore not directly comparable. Colour: split (purple system, brown energy); line style: signal (dashed AU, dotted EU, solid TU), with TU highlighted in the calibration panels; bands: 95\% intervals over five splits.}
  \label{fig:decomposition}
\end{figure}

To better understand how uncertainty components behave in different training schemes, we examine their evolution over the course of training. Figure~\ref{fig:decomposition} follows the full \LFHF\ trajectory on the model's own training data (a, b), the ID test set (c, d) and the OOD test set (e, f). On the training set, ranking changes little during DFT pretraining and improves markedly during CC fine-tuning (panel a) while the calibration of TU remains stable along the training epochs (panel b). HF-only models behave the same way on their training data (Appendix~\ref{sec:app:dynamics}).

On the ID test set, ranking performance improves for all three uncertainty signals during pretraining (panel c). The change from the low- to the high-fidelity target produces a sharp discontinuity; during the subsequent fine-tuning stage, the ranking performance of EU and TU recovers, whereas that of AU changes little. Calibration follows the opposite trend (panel d): TU becomes better calibrated during pretraining but progressively less calibrated during fine-tuning. This deterioration is driven primarily by the aleatoric component, whose calibration error increases by approximately two orders of magnitude over the fine-tuning stage. 

\begin{figure}[t!]
  \centering
  \includegraphics[width=0.98\linewidth]{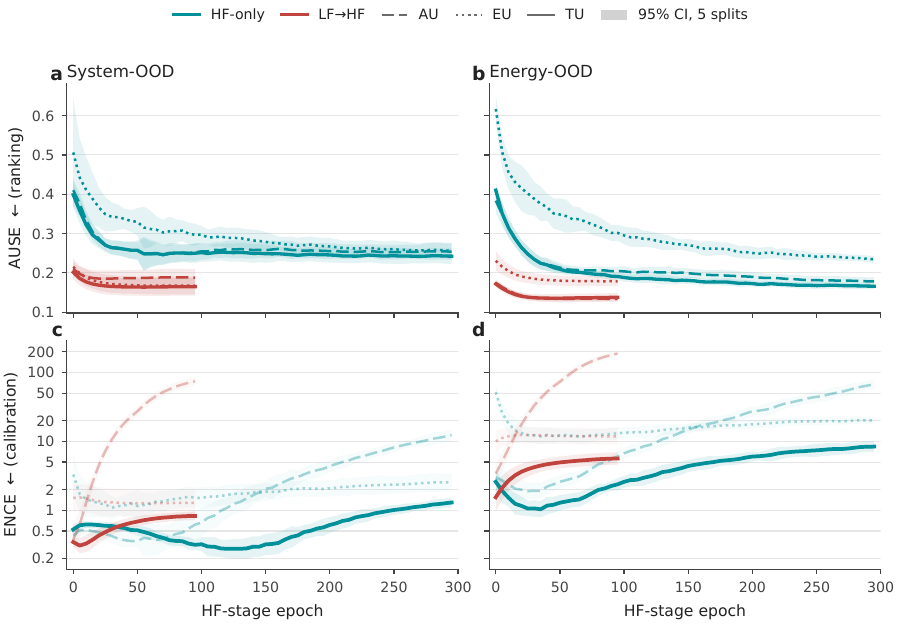}
  \caption{\textbf{Comparison of uncertainty evolution for the high-fidelity target.} AUSE (top) and ENCE (bottom, log scale) on System-OOD (a, c) and Energy-OOD (b, d) against the HF-stage epoch, for HF-only from scratch (teal, 300 epochs) and CC fine-tuning of \LFHF\ (red, 100 epochs). Line style: signal (dashed AU, dotted EU, solid TU); in the calibration panels TU is highlighted. Lines are exponential moving averages of the five-split mean (geometric for ENCE); bands are 95\% intervals. Last-epoch values differ from Figure~\ref{fig:final}, whose models are best-validation checkpoints.}
  \label{fig:hfdynamics}
\end{figure}

A similar ranking behavior is observed on the OOD test sets (panel e): ranking improves during pretraining under both distribution shifts and approaches a plateau during fine-tuning. The evolution of ENCE, however, depends on the type of shift (panel f). During pretraining, TU remains reasonably well calibrated on unseen molecular systems but progressively drifts on higher-energy configurations. Once fine-tuning begins, its calibration error increases under both shifts. However, the two uncertainty components behave differently: the calibration error associated with EU remains comparatively stable and ends near its pretraining level, whereas AU becomes strongly overconfident, with its predicted variance falling one to two orders of magnitude below the squared prediction error. The evolving balance between the two components is examined in Appendix~\ref{sec:supp:ratio}. Although the poor reliability of aleatoric uncertainty under distribution shift is well documented~\citep{valdenegro2022deeper}, its deterioration on the ID test set indicates that the effect is also associated with high-fidelity training and is accelerated by pretraining (Section~\ref{sec:results:hf}). A plausible explanation is that the variance head follows the shrinking residuals of the training configurations and transfers these small variance estimates to held-out configurations whose errors have not decreased, a known failure mode of mean--variance training~\citep{nix1994estimating,seitzer2022pitfalls}. Because the reference energies are deterministic, this behavior suggests that the aleatoric term acts less like irreducible noise and more like a reducible estimate of in-sample residual difficulty.

\subsection{Pretraining speeds up both the gain in ranking and the loss of calibration}
\label{sec:results:hf}

\begin{figure}[t!]
  \centering
  \includegraphics[width=0.95\linewidth]{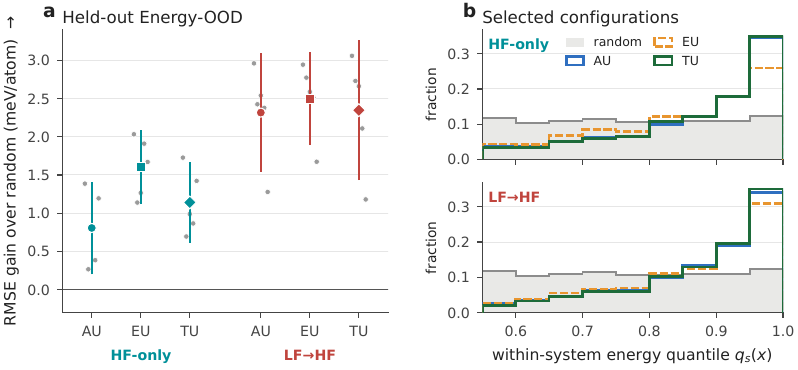}
  \caption{\textbf{Uncertainty-guided acquisition.} One round acquires 500 of 2{,}500 Energy-OOD pool configurations; models are evaluated on the disjoint held-out half. \textbf{(a)} RMSE reduction relative to random selection (positive favours uncertainty); markers are five-split means with 95\% intervals, grey dots individual splits. \textbf{(b)} Within-system energy quantile $q_s(x)$ of the acquired configurations; grey is random selection, which reproduces the pool.}
  \label{fig:acquisition}
\end{figure}

Having examined the full training dynamics of \LFHF, we now compare its high-fidelity stage with HF-only training (Figure~\ref{fig:hfdynamics}), focusing on out-of-distribution performance. Under both distribution shifts and for all three uncertainty signals, the \LFHF\ models achieve a lower AUSE than the HF-only ones already from the first epoch.  For both protocols, AUSE then decreases and levels off over the course of high-fidelity training.

Calibration follows a different trajectory (panels c and d). For HF-only training, ENCE initially decreases before increasing again, whereas for \LFHF\ it is lowest near the beginning of the high-fidelity stage and then grows with fine-tuning. As observed above, this deterioration is driven primarily by the aleatoric component. Pretraining therefore does not appear to cause the degradation, but it accelerates it considerably: within 100 fine-tuning epochs, the AU ENCE reaches about $80$ on System-OOD and $200$ on Energy-OOD, compared with about $14$ and $70$, respectively, after 300 epochs of HF-only training. The epistemic component behaves differently: its calibration error ends close to or below its initial value in all four settings, limiting the deterioration of total uncertainty relative to that of the aleatoric component. Overall, these results show that uncertainty ranking and calibration follow distinct, partially decoupled trajectories during training: ranking can continue to improve even as calibration deteriorates. Checkpoint selection should therefore not rely on prediction error alone, but should jointly monitor accuracy, uncertainty ranking, and calibration throughout training.

\subsection{Pretrained uncertainty estimates improve single-round acquisition}
\label{sec:results:al}

Finally, we compare single- and multi-fidelity training in a setting where uncertainty is used to acquire new data. We perform a single acquisition round on Energy-OOD, selecting 500 configurations either at random or according to the highest AU, EU, or TU scores. After adding the corresponding high-fidelity labels to the training set, each model continues training and is evaluated on a disjoint held-out set (Appendix~\ref{sec:app:al}). All uncertainty-based acquisition strategies outperform random selection (Figure~\ref{fig:acquisition}a). However, all three uncertainty signals yield larger improvements for the pretrained ensembles than for those trained only on high-fidelity data: relative to random selection, their RMSE decreases by about $2.3$--$2.5$~meV/atom, compared with $0.8$--$1.6$~meV/atom for HF-only training. This is notable because the pretrained ensemble is already more accurate before acquisition, with an RMSE of about $59$~meV/atom on the held-out set compared with $75$~meV/atom for the HF-only ensemble, yet it benefits more from uncertainty-guided selection. The two protocols select configurations with similar energy-quantile distributions (Figure~\ref{fig:acquisition}b), indicating that the larger improvement after pretraining cannot be explained simply by selecting systematically higher-energy configurations. Instead, it suggests that pretraining produces a more informative uncertainty-based acquisition function within comparable energy ranges.

\section{Conclusion}
\label{sec:conclusions}
We studied how low-fidelity pretraining followed by high-fidelity fine-tuning shapes the ranking, calibration, and decomposition of predictive uncertainty in atomistic models.
Low-fidelity pretraining improves predictive accuracy and the error ranking for every uncertainty signal across all test sets, but does not consistently improve total uncertainty calibration. It also changes the uncertainty decomposition: without pretraining, the epistemic component is the weakest ranking signal; after pretraining it approaches total uncertainty on three of the four test sets (\textbf{RQ1}). Under both shifts, errors are ranked better out-of-domain than in-domain, whereas only energy extrapolation degrades calibration (\textbf{RQ2}). Along training, most calibration loss arises during the high-fidelity stage under both protocols, and develops faster after pretraining: aleatoric uncertainty remains conservative on the training data but becomes strongly overconfident on unseen configurations, while epistemic uncertainty mitigates the degradation of total uncertainty (\textbf{RQ3}). Finally, pretrained ensembles yield more effective uncertainty-based acquisition functions than HF-only models (\textbf{RQ4}).

These findings lead to clear practical implications. Ranking and calibration should be evaluated separately, and total uncertainty calibration should be monitored on held-out data throughout training because it may deteriorate even as ranking improves. Checkpoint selection should therefore account for uncertainty ranking and calibration in addition to prediction error.  Aleatoric estimates obtained after training on scarce high-fidelity data require particular caution on unseen configurations, and in-domain recalibration should be validated explicitly before use under extrapolation (Appendix~\ref{sec:supp:recal}). Although our analysis is limited to energy prediction, one uncertainty estimator, and a single acquisition setting (Appendix~\ref{sec:limitations}), it establishes a broader conclusion: uncertainty in multi-fidelity atomistic models is not a static property, with its ranking, calibration, and decomposition changing across fidelity transitions, distribution shifts, and training stages.

\subsection*{AI use statement}

In this work, we used generative AI tools (large language model coding assistants) to help write and refactor analysis and plotting code and to draft and edit parts of the manuscript text. We have not used generative AI tools to generate research ideas, experimental data or results. All AI-assisted code was reviewed by the authors, and every number reported in the paper was recomputed from the experiment outputs by scripts in the released repository. We have reviewed all AI-assisted work and take responsibility for the final content of this work, including text, claims and artifacts produced with the aid of generative AI.

\subsection*{Ethics statement}

This work uses only public data sets of molecular configurations labelled with electronic-structure methods; it involves no human subjects or personal data. Possible misuse and the broader impact of the work are discussed in Appendix~\ref{sec:broader_impact}.

\subsection*{Reproducibility statement}

All data are public: ANI-1x~\citep{smith2020ani}. Code, the exact data splits, configuration files, evaluation scripts and the scripts that produce every figure are available at the repository linked in Section~\ref{sec:intro}. Section~\ref{sec:experiments} summarizes the data, shifts, training protocols, checkpoint selection and statistical protocol. Appendix~\ref{sec:app:theory} states the assumptions behind the estimator and the evaluation, Appendices~\ref{sec:system_split} and~\ref{sec:energy_split} detail the two splitting procedures, Appendix~\ref{sec:hyperparameters} the model, the optimization, the random seeds, the selected checkpoints and the computational resources, Appendix~\ref{sec:app:filter} the training-support filter, Appendix~\ref{sec:uq_metrics_appendix} the metrics, Appendix~\ref{sec:supp:recal} the recalibration and Appendix~\ref{sec:app:al} the acquisition protocol.

\bibliography{references}
\bibliographystyle{plainnat}

\appendix
\renewcommand{\topfraction}{0.95}\renewcommand{\textfraction}{0.05}\renewcommand{\floatpagefraction}{0.85}

\section{Related Work}
\label{sec:related_work}

Recent surveys and perspectives provide broad overviews of the uncertainty quantification (UQ) field, covering Bayesian methods, ensembles, calibration strategies, active learning, and the relationship between uncertainty, dataset composition, transferability, and model robustness~\citep{dai2025uncertainty,grasselli2025uncertainty,he2026survey}. These works highlight that benchmarking and interpretation of UQ methods remain open challenges, especially in transfer and multi-fidelity settings.

A large body of work has focused on practical uncertainty estimators for neural network potentials. Ensemble-based methods remain among the most widely used and robust approaches. For instance, \citet{tan2023single} compare ensembles against single-model uncertainty methods such as mean-variance estimation, evidential regression, and Gaussian mixture models, finding that no method uniformly dominates across datasets and metrics, while ensembles remain strong for robustness and generalization. Other work has proposed more efficient ensemble variants, including multi-headed architectures, committee models, shallow or last-layer ensembles, and shared-representation ensembles, to reduce computational cost while retaining useful uncertainty estimates~\citep{carrete2023deep,kellner2024uncertainty,beck2025multi}. 

Several works have specifically addressed calibration and uncertainty decomposition in atomistic models. \citet{busk2023graph} propose a framework for calibrated aleatoric and epistemic uncertainty in graph neural network potentials, using deep ensembles to decompose total uncertainty and post-hoc recalibration to improve agreement between predicted uncertainty and observed error. More generally, \citet{valdenegro2022deeper} study disentangled aleatoric and epistemic uncertainty across different UQ methods, showing that the interaction between these components can be non-trivial, that aleatoric uncertainty may become unreliable under distribution shift, and that ensembles provide the most consistent decomposition. In molecular property prediction, \citet{heid2023characterizing} further analyze uncertainty sources by separating epistemic uncertainty into bias and variance contributions, showing that different sources of error dominate depending on noise level, dataset size, model architecture, and representation.

A related line of work studies multi-fidelity learning in atomistic modeling. \citet{gardner2025understanding} investigate multi-fidelity training strategies for atomistic models, including pretraining--fine-tuning and multi-headed training, showing that fine-tuning can improve accuracy but that learned representations may remain method-specific, while multi-headed models can learn more method-agnostic representations across fidelities. More broadly, multi-fidelity UQ has been studied in surrogate modeling and materials design, often using Gaussian process models to combine information across fidelity levels while quantifying predictive confidence~\citep{tran2020multi}.

Finally, the interpretation of uncertainty depends strongly on the evaluation metric. \citet{rasmussen2023uncertain,lind2024uncertainty} analyze common regression UQ metrics, including AUSE, calibration error, Spearman correlation, and NLL, showing that they capture different properties and can lead to different conclusions.

Overall, existing work has made substantial progress on uncertainty estimation for atomistic models, uncertainty decomposition, calibration, efficient ensemble methods, and multi-fidelity training. However, these directions have largely been studied separately. In contrast, our work provides a controlled empirical analysis of aleatoric and epistemic uncertainty across multi-fidelity training regimes, system-level and energy-based distribution shifts, and training dynamics. This allows us to study not only whether uncertainty is accurate at convergence, but also how its components evolve and whether they can serve as diagnostic signals for model behavior under changing fidelity and data regimes.

\section{Technical background: error decomposition and mean--variance ensembles}
\label{sec:app:theory}

This appendix collects standard results behind the estimator of Section~\ref{sec:predictive_model} and the way we evaluate it. None of them is new; we state them in our notation to make explicit what AU, EU and TU can and cannot measure when reference energies are deterministic. All identities below are homogeneous of degree two in the energies, so they hold unchanged for the per-atom quantities used throughout the paper.

\paragraph{Bias--variance decomposition.} Let $y(\mathbf{x})$ be the reference energy of configuration $\mathbf{x}$, deterministic for a given electronic-structure method, and let $f(\mathbf{x};\mathcal{D},\omega)$ be a model trained on data $\mathcal{D}$ with random seed $\omega$ (initialization and data order). With $\bar f(\mathbf{x})=\mathbb{E}_{\mathcal{D},\omega}[f(\mathbf{x};\mathcal{D},\omega)]$, the expected squared error decomposes as~\citep{geman1992neural}
\begin{equation}
\label{eq:biasvar}
\mathbb{E}_{\mathcal{D},\omega}\!\left[\big(f(\mathbf{x};\mathcal{D},\omega)-y(\mathbf{x})\big)^2\right]
=\underbrace{\big(\bar f(\mathbf{x})-y(\mathbf{x})\big)^2}_{\text{bias}^2}
+\underbrace{\mathbb{E}_{\mathcal{D},\omega}\!\left[\big(f(\mathbf{x};\mathcal{D},\omega)-\bar f(\mathbf{x})\big)^2\right]}_{\text{variance}}.
\end{equation}
With noisy labels, an irreducible noise term would add to the right-hand side; with deterministic labels it vanishes, so the whole expected error is reducible in principle, the variance by more data and the bias by a better model class. A deep ensemble whose members share the training set and differ only in their seed samples the variation over $\omega$ at fixed $\mathcal{D}$, which is substantial for neural networks~\citep{fort2019deep}. By the law of total variance, its expectation over data sets is at most the variance term of Equation~\ref{eq:biasvar}, and it is blind to the bias, which all members share. EU is therefore an operational and generally partial estimate of reducible error.

\paragraph{Mean--variance estimation.} Each member is trained with the Gaussian NLL of Section~\ref{sec:predictive_model}. For a fixed mean $\mu(\mathbf{x})$, the expected loss under a target distribution $p(y\mid\mathbf{x})$ is minimized by the conditional mean squared residual~\citep{nix1994estimating},
\begin{equation}
\sigma^{2\star}(\mathbf{x})=\mathbb{E}\big[(y-\mu(\mathbf{x}))^2\,\big|\,\mathbf{x}\big].
\end{equation}
With deterministic labels, $p(y\mid\mathbf{x})$ is a point mass, and a variance head fitted to finitely many configurations approximates a local average of its member's squared training residuals, whatever their origin (limited data, misspecification or optimization), rather than observation noise. Two known properties of the estimator matter for our results~\citep{skafte2019reliable,seitzer2022pitfalls,stirn2023faithful,sluijterman2024optimal}. First, the head is fitted to training residuals, which shrink as the network fits its training set, so on unseen configurations it can inherit the scale of the training error rather than that of the test error; this is the behavior of AU observed in Sections~\ref{sec:results:decomposition} and~\ref{sec:results:hf}. Second, the gradient with respect to the mean is weighted by $1/\sigma^2$, so configurations that the head already deems easy dominate the fit of the mean~\citep{seitzer2022pitfalls}. We use the plain Gaussian NLL, without the modifications proposed in these works.

\paragraph{Ensembles and the law of total variance.} Treating the $M$ members as samples from an approximate posterior, the predictive distribution is the mixture $\frac{1}{M}\sum_{m}\mathcal{N}\big(\hat E^{(m)}(\mathbf{x}),\hat\sigma^{2,(m)}(\mathbf{x})\big)$. Its mean is the ensemble mean $\bar E(\mathbf{x})$ and its variance is exactly AU\,+\,EU in Equation~\ref{eq:tu}, the sample counterpart of Equation~\ref{eq:total_variance}~\citep{lakshminarayanan2017simple,kendall2017uncertainties,depeweg2018decomposition}. TU is therefore the variance of the predictive distribution around the prediction we evaluate, and the only one of the three signals that the observed error of $\bar E$ can calibrate. The split into AU and EU is operational~\citep{hullermeier2021aleatoric}: neither term has an observable target of its own against the error of the ensemble mean, so their ENCE in the main text describes how much of that error each term accounts for, not whether the term is calibrated.

\paragraph{The variance heads and the error of the ensemble mean.} The members themselves do provide observable targets, since each head is trained to match its own member's squared error. Let $e_m=y-\hat E^{(m)}(\mathbf{x})$ be the error of member $m$ and $e=y-\bar E(\mathbf{x})$ that of the ensemble mean. Because the member deviations $\hat E^{(m)}-\bar E$ average to zero, the ambiguity decomposition~\citep{krogh1994neural} holds exactly for every configuration,
\begin{equation}
\label{eq:ambiguity}
e^2=\frac{1}{M}\sum_{m=1}^{M}e_m^2-\mathrm{EU},
\qquad\text{hence}\qquad
\mathrm{TU}-e^2=\Big(\mathrm{AU}-\frac{1}{M}\sum_{m=1}^{M}e_m^2\Big)+2\,\mathrm{EU},
\end{equation}
where EU is the population variance of the member predictions, as in Equation~\ref{eq:tu}. The bracket compares each head with the quantity it is trained to match, and, being linear, the identity also holds for the bin averages used by ENCE. Two consequences follow. If every head matched its member's squared error, TU would exceed the squared error of the ensemble mean by $2\,\mathrm{EU}$, that is, it would be calibrated or conservative; this is consistent with TU lying on or slightly below the diagonal in domain (Appendix~\ref{sec:supp:reliability}). Conversely, TU can underestimate the error of the ensemble mean only where the heads underestimate their members' errors by more than $2\,\mathrm{EU}$. The overconfidence of TU on unseen data (Sections~\ref{sec:results:decomposition} and~\ref{sec:results:hf}) is therefore a failure of the variance heads, a conclusion that needs no separate target for AU. Finally, the average member error exceeds $e^2$ by EU, which is typically smaller than AU (Appendix~\ref{sec:supp:ratio}); where AU falls far short of the error, as on unseen data after fine-tuning, comparing AU with $e^2$ rather than with the members' own errors therefore changes little.

\section{Details on uncertainty evaluation metrics}
\label{sec:uq_metrics_appendix}

In this section, we provide the formal definitions of the metrics used in the main text. These metrics are designed to capture complementary aspects of predictive uncertainty, including ranking quality, calibration, and probabilistic correctness.

\subsection{Area under the sparsification error (AUSE)}

AUSE evaluates how well uncertainty ranks samples according to their prediction errors.

Let $S = \{(x_i, y_i)\}_{i=1}^{N}$ be a test set, and define the mean absolute error:

\begin{equation}
\mathrm{MAE}(S) = \frac{1}{|S|} \sum_{(x_i,y_i)\in S} |y_i - \hat{y}(x_i)|.
\end{equation}

For a fraction $\alpha \in [0,1]$, let:
\begin{itemize}
    \item $S_V(\alpha)$ denote the subset obtained by removing the top-$\alpha$ fraction of samples with the largest prediction errors (oracle ranking),
    \item $S_U(\alpha)$ denote the subset obtained by removing the top-$\alpha$ fraction of samples with the largest predicted uncertainties.
\end{itemize}

The sparsification error compares the degradation in performance when removing samples based on uncertainty versus the oracle ranking. AUSE is defined as:

\begin{equation}
\mathrm{AUSE}(S) =
\frac{1}{\mathrm{MAE}(S)}
\int_0^1
\left(
\mathrm{MAE}(S_U(\alpha))
-
\mathrm{MAE}(S_V(\alpha))
\right)
\, d\alpha.
\end{equation}

AUSE measures the \emph{ranking quality} of uncertainty. Low AUSE indicates that uncertainty successfully identifies high-error samples. Compared to pure ordering metrics such as Spearman correlation, AUSE is sensitive to the \emph{magnitude} of errors, penalizing misranking of large-error samples more strongly.

\subsection{Spearman rank correlation}

Spearman's rank correlation evaluates the monotonic relationship between uncertainty and prediction error:

\begin{equation}
\rho_S = \mathrm{Corr}_{\mathrm{rank}}
\big(
\sigma^2(x),\, e^2(x)
\big),
\end{equation}

where $e(x) = y - \hat{y}(x)$.

This metric measures whether uncertainty preserves the ordering of errors. It is invariant to monotonic transformations of the uncertainty scale (e.g., post-hoc calibration), and therefore isolates the \emph{informativeness} of uncertainty independently of its magnitude.

\subsection{Binned expected normalized calibration error (ENCE)}

Calibration evaluates whether predicted uncertainty magnitudes match empirical errors.

We partition the test set into $B$ bins based on predicted uncertainty. For each bin $b$, we compute:

\begin{equation}
\mathrm{RMSE}_b = \sqrt{\mathbb{E}[e^2 | b]},
\qquad
\mathrm{RMV}_b = \sqrt{\mathbb{E}[\sigma^2 | b]}.
\end{equation}

The Expected Normalized Calibration Error (ENCE) is defined as:

\begin{equation}
\mathrm{ENCE} = \frac{1}{B}
\sum_{b=1}^{B}
\frac{|\mathrm{RMV}_b^2 - \mathrm{RMSE}_b^2|}
{\mathrm{RMV}_b^2}.
\end{equation}

ENCE measures the \emph{scale correctness} of uncertainty. Unlike AUSE and Spearman, it does not assess ranking quality, but instead quantifies whether predicted variances match observed squared errors.

We use $B=15$ bins with an equal number of configurations each.

\subsection{Negative log-likelihood (NLL)}

NLL evaluates the probabilistic correctness of the predictive distribution:

\begin{equation}
\mathrm{NLL}(S)
=
-\frac{1}{N}
\sum_{i=1}^{N}
\log p(y_i | x_i).
\end{equation}

Under the Gaussian predictive assumption:

\begin{equation}
\mathrm{NLL}
=
\frac{1}{2}
\left(
\frac{e^2}{\sigma^2}
+
\log \sigma^2
\right)
+ C,
\end{equation}

where $C$ is a constant.

In our evaluation, $e$ is the per-atom error of the ensemble mean and $\sigma^2$ the per-atom TU, i.e.\ the Gaussian that matches the first two moments of the ensemble mixture (Appendix~\ref{sec:app:theory}); the NLL, including the constant $\tfrac{1}{2}\log 2\pi$, is averaged over the configurations of a test set. NLL is a strictly proper scoring rule and jointly evaluates calibration and sharpness. However, it does not isolate ranking behavior and can be dominated by a small number of poorly calibrated predictions.

\section{System-level data splitting protocol}
\label{sec:system_split}

\paragraph{Dataset structure.}
The ANI-1x dataset consists of 3{,}114 molecular systems (distinct chemical formulas), each associated with a variable number of molecular configurations (geometries). For each configuration, multiple quantum-chemical annotations may be available, including density functional theory (DFT) energies and, for a subset, high-fidelity coupled-cluster energies (CCSD(T)/CBS).

\subsection{System-OOD split}

Let $\mathcal{S}$ denote the set of molecular systems. We first perform a random partition of $\mathcal{S}$ into two disjoint subsets:
\begin{equation}
\mathcal{S} = \mathcal{S}_{\text{seen}} \cup \mathcal{S}_{\text{OOD}},
\qquad
\mathcal{S}_{\text{seen}} \cap \mathcal{S}_{\text{OOD}} = \varnothing,
\end{equation}
where $\mathcal{S}_{\text{OOD}}$ contains a fixed proportion ($40\%$) of systems and is reserved exclusively for \emph{System-OOD testing}. All configurations belonging to systems in $\mathcal{S}_{\text{OOD}}$ are excluded from training and validation.

\subsection{In-domain (ID) split within seen systems}

The remaining systems $\mathcal{S}_{\text{seen}}$ define the in-domain (ID) regime. For each system $s \in \mathcal{S}_{\text{seen}}$, we randomly shuffle its configurations and partition them into three disjoint subsets according to fixed proportions:
\begin{equation}
\mathcal{X}_s = 
\mathcal{X}_s^{\text{train}} 
\cup 
\mathcal{X}_s^{\text{val}} 
\cup 
\mathcal{X}_s^{\text{test-ID}},
\end{equation}
where $\mathcal{X}_s^{\text{train}}$, $\mathcal{X}_s^{\text{val}}$, and $\mathcal{X}_s^{\text{test-ID}}$ denote the training, validation, and in-domain test configurations of system $s$, respectively.

Aggregating across systems yields:
\begin{align}
\mathcal{X}^{\text{train}}_{\text{ID}} &= \bigcup_{s \in \mathcal{S}_{\text{seen}}} \mathcal{X}_s^{\text{train}}, \\
\mathcal{X}^{\text{val}}_{\text{ID}}   &= \bigcup_{s \in \mathcal{S}_{\text{seen}}} \mathcal{X}_s^{\text{val}}, \\
\mathcal{X}^{\text{test}}_{\text{ID}}  &= \bigcup_{s \in \mathcal{S}_{\text{seen}}} \mathcal{X}_s^{\text{test-ID}}.
\end{align}

This procedure ensures that:
(i) no configuration leakage occurs across splits, and  
(ii) the in-domain test set contains unseen configurations from the same distribution of molecular systems as training.

\subsection{Annotation-specific dataset construction}

From each configuration split, we construct task-specific datasets:

\begin{itemize}
    \item \textbf{DFT datasets:} configurations with available DFT annotations are selected from each split, optionally using stratified sampling to prevent over-representation of large systems.
    \item \textbf{CC datasets:} high-fidelity CC configurations are selected \emph{only from the already selected DFT configurations}, ensuring
    \begin{equation}
        \mathcal{X}_{\text{CC}} \subseteq \mathcal{X}_{\text{DFT}},
    \end{equation}
    which guarantees consistency for multi-fidelity pretraining and fine-tuning.
\end{itemize}

\subsection{Stratified sampling across molecular systems}
\label{sec:stratified_sampling}

After defining the system-level splits, we optionally subsample configurations within each split to reach fixed target sizes (e.g., $n^{\text{train}}_{\text{DFT}}$, $n^{\text{val}}_{\text{DFT}}$). 
To avoid over-representation of large systems (some of which contain orders of magnitude more configurations than others) we employ a \emph{stratified sampling scheme at the system level}.

\paragraph{Available pools.}
For a given split $\sigma \in \{\text{train}, \text{val}, \text{test-ID}, \text{test-OOD}\}$ and theory level (DFT or CC), let
\begin{equation}
    \mathcal{X}_{s,\sigma}
\end{equation}
denote the set of configurations of system $s$ that:
(i) belong to split $\sigma$, and 
(ii) have a valid annotation for the selected theory level.
Let
\begin{equation}
    n_{s,\sigma} = |\mathcal{X}_{s,\sigma}|
\end{equation}
be the number of available candidates for system $s$ in split $\sigma$.
The total number of available configurations in that split is
\begin{equation}
    N_{\sigma} = \sum_{s} n_{s,\sigma}.
\end{equation}

\paragraph{Quota allocation.}
Given a desired target size $N_{\sigma}^{\text{target}} \le N_{\sigma}$, we allocate per-system quotas $q_{s,\sigma}$ proportionally to the number of available configurations:
\begin{equation}
    \tilde{q}_{s,\sigma}
    =
    N_{\sigma}^{\text{target}}
    \frac{w_{s,\sigma}}{\sum_{s'} w_{s',\sigma}},
\end{equation}
where the weights are defined as
\begin{equation}
    w_{s,\sigma} =
    \begin{cases}
        n_{s,\sigma}, & \text{if no cap is applied}, \\
        \min(n_{s,\sigma}, \kappa), & \text{with optional cap } \kappa.
    \end{cases}
\end{equation}
The optional cap $\kappa$ limits the influence of very large systems while still allowing them to contribute more samples when necessary.

The final integer quotas $q_{s,\sigma}$ are obtained by:
\begin{enumerate}
    \item taking the floor of $\tilde{q}_{s,\sigma}$, and
    \item distributing the remaining samples to systems with the largest fractional remainders,
\end{enumerate}
ensuring that
\begin{equation}
    \sum_s q_{s,\sigma} = N_{\sigma}^{\text{target}}.
\end{equation}

\paragraph{Sampling without replacement.}
For each system $s$, we then draw $q_{s,\sigma}$ configurations uniformly at random without replacement from $\mathcal{X}_{s,\sigma}$. 
This yields the final stratified subset
\begin{equation}
    \widehat{\mathcal{X}}_{\sigma}
    =
    \bigcup_s \widehat{\mathcal{X}}_{s,\sigma},
\end{equation}
with $|\widehat{\mathcal{X}}_{\sigma}| = N_{\sigma}^{\text{target}}$.

\paragraph{Multi-fidelity constraint.}
For the high-fidelity CC dataset, candidate configurations are first restricted to those already selected in the corresponding DFT split, enforcing
\begin{equation}
    \mathcal{X}^{\text{CC}}_{\sigma}
    \subseteq
    \widehat{\mathcal{X}}^{\text{DFT}}_{\sigma}.
\end{equation}
Stratified quota allocation is then applied within this restricted pool, guaranteeing consistency between low- and high-fidelity datasets.

\paragraph{Properties.}
This procedure:
\begin{itemize}
    \item preserves the system-level split,
    \item prevents dominance of large systems,
    \item enforces exact target sizes,
    \item maintains $\text{CC} \subseteq \text{DFT}$,
    \item and is reproducible given the split seed (we use five seeds, $0$--$4$).
\end{itemize}

\subsection{Resulting evaluation regimes}

This protocol produces four disjoint evaluation settings:

\begin{itemize}
    \item \textbf{Train (ID)}: configurations from $\mathcal{S}_{\text{seen}}$ used for model optimization.
    \item \textbf{Validation (ID)}: configurations from $\mathcal{S}_{\text{seen}}$ used for checkpoint selection and for the post-hoc recalibration of Appendix~\ref{sec:supp:recal}.
    \item \textbf{Test-ID}: unseen configurations from $\mathcal{S}_{\text{seen}}$.
    \item \textbf{Test-OOD (System-OOD)}: configurations from $\mathcal{S}_{\text{OOD}}$, representing novel molecular systems.
\end{itemize}

By construction, the System-OOD evaluation measures generalization to entirely unseen molecular systems, while the Test-ID split quantifies generalization within the same system distribution without configuration leakage. This split design enables a controlled comparison between in-domain performance and true structural distributional shift.

\section{Energy-based data splitting protocol}
\label{sec:energy_split}

To evaluate generalization across energetic regimes within each molecular system, we construct an energy-based split that separates low-energy (in-domain) configurations from high-energy (out-of-domain) configurations.

\paragraph{Per-system relative energies.}
For each molecular system $s$, let $\mathcal{X}_s$ denote its full set of configurations.
Given an energy annotation $E_s(x)$ (by default the DFT energy), we compute relative energies
\begin{equation}
\Delta E_s(x) = E_s(x) - \min_{x' \in \mathcal{X}_s} E_s(x').
\end{equation}
Relative energies are computed independently for each system to remove absolute offsets.

\paragraph{Empirical energy quantiles.}
Let $\{x_i\}_{i=1}^{N_s}$ be the configurations of system $s$ with valid energies.
We assign to each configuration an empirical quantile rank
\begin{equation}
q_s(x_i) = \frac{\mathrm{rank}(\Delta E_s(x_i))}{N_s - 1},
\end{equation}
where ranks are computed after sorting $\Delta E_s$ in ascending order.
This yields $q_s(x) \in [0,1]$ for all valid configurations.

\paragraph{Energy domains.}
Given two thresholds $q_{\text{low}} < q_{\text{high}}$ (we use $0.5$ and $0.55$), we define:
\begin{align}
\mathcal{X}_s^{\text{low}} &= \{ x \in \mathcal{X}_s : q_s(x) \le q_{\text{low}} \}, \\
\mathcal{X}_s^{\text{high}} &= \{ x \in \mathcal{X}_s : q_s(x) \ge q_{\text{high}} \}.
\end{align}
Configurations with $q_{\text{low}} < q_s(x) < q_{\text{high}}$ form an intermediate band that is excluded from all splits to ensure a clear separation between in-domain and out-of-domain samples.

\paragraph{Low-energy ID splits.}
For each system, the low-energy set $\mathcal{X}_s^{\text{low}}$ is randomly shuffled and partitioned into:
\begin{equation}
\mathcal{X}_s^{\text{low}}
=
\mathcal{X}_s^{\text{train}}
\cup
\mathcal{X}_s^{\text{val}}
\cup
\mathcal{X}_s^{\text{test-ID}},
\end{equation}
according to fixed proportions $(p_{\text{train}}, p_{\text{val}}, p_{\text{test-ID}})$.
Aggregating across systems yields:
\begin{align}
\mathcal{X}^{\text{train}}_{\text{ID}} &= \bigcup_s \mathcal{X}_s^{\text{train}}, \\
\mathcal{X}^{\text{val}}_{\text{ID}} &= \bigcup_s \mathcal{X}_s^{\text{val}}, \\
\mathcal{X}^{\text{test-ID}}_{\text{ID}} &= \bigcup_s \mathcal{X}_s^{\text{test-ID}}.
\end{align}

\paragraph{Energy-OOD set.}
All high-energy configurations are assigned to the OOD pool:
\begin{equation}
\mathcal{X}^{\text{test-OOD}}
=
\bigcup_s \mathcal{X}_s^{\text{high}}.
\end{equation}

\paragraph{Theory-specific datasets.}
For each split $\sigma \in \{\text{train}, \text{val}, \text{test-ID}, \text{test-OOD}\}$:
\begin{itemize}
    \item DFT configurations are selected from $\mathcal{X}_\sigma$ using stratified sampling across systems.
    \item CC configurations are selected exclusively from the corresponding DFT subset, enforcing
    \begin{equation}
        \mathcal{X}^{\text{CC}}_\sigma \subseteq \mathcal{X}^{\text{DFT}}_\sigma.
    \end{equation}
\end{itemize}

\paragraph{Properties.}
This protocol:
\begin{itemize}
    \item defines energy shift independently within each molecular system,
    \item avoids configuration leakage across splits,
    \item enforces a strict low/high separation,
    \item preserves per-system diversity via stratified sampling,
    \item guarantees $\text{CC} \subseteq \text{DFT}$,
    \item and is reproducible given the split seed (we use five seeds, $0$--$4$).
\end{itemize}

\section{Training details and hyperparameters}
\label{sec:hyperparameters}

\paragraph{Data.} Low-fidelity labels are the ANI-1x $\omega$B97x/def2-TZVPP energies (\texttt{wb97x\_tz.energy}) and high-fidelity labels the CCSD(T)/CBS energies (\texttt{ccsd(t)\_cbs.energy}). Forces are not used at either fidelity. Table~\ref{tab:sizes} lists the number of configurations per split; each CC set is a subset of the corresponding DFT set. The system split holds out $40\%$ of the molecular systems ($p_{\text{seen}}=0.6$) and divides the configurations of seen systems into train, validation and test-ID with proportions $0.6/0.2/0.2$. The energy split uses $q_{\text{low}}=0.5$ and $q_{\text{high}}=0.55$ with the same proportions. Stratified sampling uses a per-system soft cap $\kappa=64$. The whole construction is repeated with five split seeds ($0$--$4$).

\begin{table}[h]
\caption{Number of configurations per split for the DFT and CC data sets, identical for both shifts and all five split seeds.}
\label{tab:sizes}
\begin{center}
\begin{tabular}{lcc}
\toprule
Split & DFT & CC \\
\midrule
Train      & 50{,}000 & 5{,}000 \\
Validation & 10{,}000 & 1{,}000 \\
Test-ID    & 50{,}000 & 5{,}000 \\
Test-OOD   & 50{,}000 & 5{,}000 \\
\bottomrule
\end{tabular}
\end{center}
\end{table}

\paragraph{Model.} All models are MACE networks with 128 channels, $L_{\max}=1$ equivariant features, a $6$~\AA\ cutoff, two interaction layers of correlation order three and spherical harmonics up to $\ell=3$, with the mean--variance readout of Section~\ref{sec:predictive_model} and atomic reference energies fitted to the training set by least squares. The readout outputs, for every atom, an energy and an unconstrained scalar that a softplus maps to a positive variance, plus $10^{-12}$ for numerical stability; the per-atom variances are summed over the configuration and multiplied by the square of the model's energy scale factor, while the reference energies carry no variance. Each ensemble has $M=10$ members that differ only in their random seed ($0$--$9$).

\paragraph{Optimization.} Models are trained on the Gaussian NLL of the energy, with the term of each configuration divided by its number of atoms, using Adam (AMSGrad), batch size 64, weight decay $5\cdot10^{-7}$ and a reduce-on-plateau learning-rate schedule (factor $0.8$, patience 50 epochs), without stochastic weight averaging or an exponential moving average of the weights. LF-only and HF-only models are trained from scratch for 300 epochs with learning rate $10^{-2}$. \LFHF\ models are initialized from the weights of the LF-only member with the same split and seed after its last pretraining epoch, including the mean--variance readout, and fine-tuned on CC energies for 100 epochs with learning rate $10^{-3}$; the atomic reference energies are refitted on the CC training set, and the optimizer and learning-rate schedule start afresh. For every member, the checkpoint with the lowest validation NLL is kept; test sets are never used for model selection. Across the 50 members of each experiment, the selected epoch has a median of 60 (range 30--111) for HF-only under system shift and 67 (34--111) under energy shift, and a median of 2 (range 1--3) fine-tuning epochs for \LFHF\ under both shifts.

\subsection{Computational resources}
\label{sec:comp_res}

All experiments were conducted on a high-performance computing node equipped with two AMD EPYC 7742 CPUs (128 cores total, 256 threads) with 1TB of total RAM, and 8 NVIDIA A100 GPUs (40GB memory each).

Model training and evaluation were performed using GPU acceleration. Training the 450 ensemble members reported in the paper took about 1{,}900 GPU hours in total, dominated by the 100 low-fidelity pretraining runs, each of which takes about 15--16 hours on one A100 GPU and occupies about 20GB of GPU memory; high-fidelity training from scratch takes 1--1.5 hours and fine-tuning about half an hour. Per-epoch evaluation and the acquisition experiment add to this budget. The use of deep ensembles, involving multiple independently initialized models, constitutes the primary computational cost of the study.

Data preprocessing and auxiliary computations were performed on CPU resources. 

The full research project required more computing than the experiments reported in the paper due to preliminary or failed experiments.

\section{Training-support filter}
\label{sec:app:filter}

Energy-OOD contains a small number of configurations compressed far below the repulsive wall covered by training. For each split we compute the smallest interatomic distance occurring anywhere in that split's DFT and CC training and validation sets, and we exclude every test configuration whose smallest interatomic distance falls below it. The rule reads geometry only: it never inspects predictions, errors or uncertainties, and it removes the same configurations for every protocol, signal and metric within a split. It removes 13--16 of the 5{,}000 Energy-OOD configurations per split, at most two Energy-ID configurations, and none on the system split, whose training data already cover compressed geometries.

The filter matters because all our metrics average squared errors or variances. On the excluded configurations the models extrapolate very poorly, and a single one of them can dominate the RMSE, the ENCE of a bin or the AUSE of a whole test set. We therefore treat them as outside the domain of validity of the models rather than as evidence about the quality of uncertainty. The same rule is applied to the acquisition experiment (Appendix~\ref{sec:app:al}).

\section{Training dynamics}
\label{sec:app:dynamics}

\begin{figure}[tbp]
  \centering
  \includegraphics[width=5.5in]{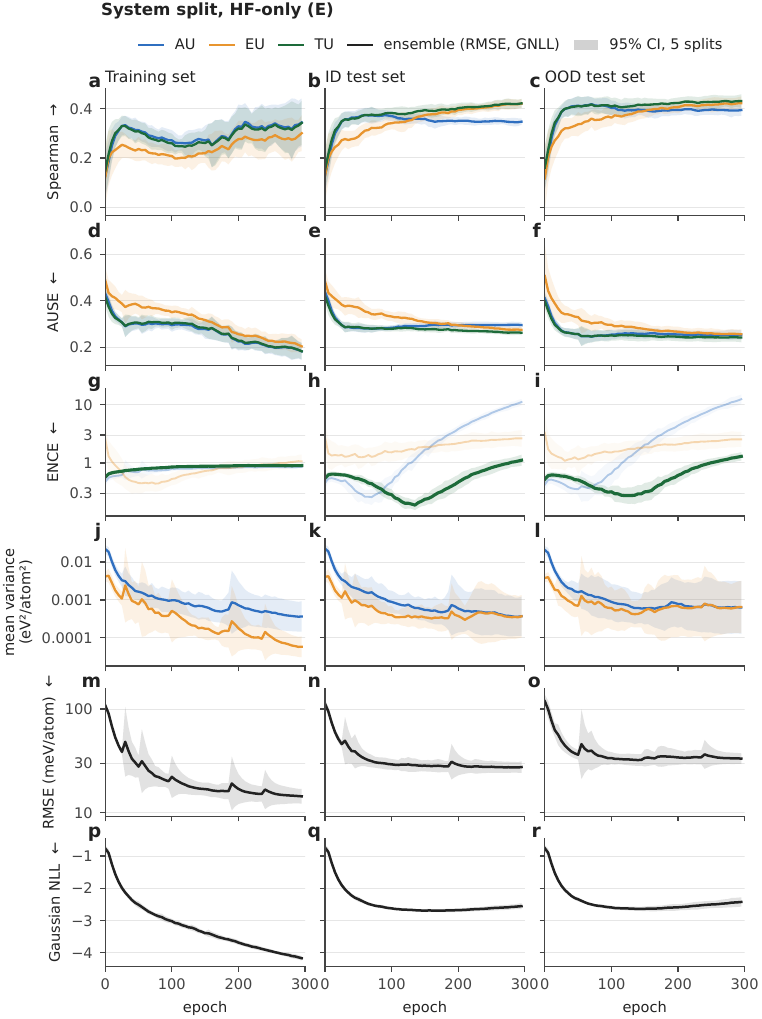}
  \caption{\textbf{Training dynamics of HF-only ensembles under system shift.} Columns: training set, ID and OOD test set. Rows: Spearman ($\uparrow$), AUSE ($\downarrow$), ENCE ($\downarrow$, TU highlighted, AU and EU faded), mean per-atom AU and EU, RMSE of the ensemble mean and Gaussian NLL of the predictive distribution, against epoch. Lines are exponential moving averages of the five-split means (geometric on log axes); bands are 95\% intervals.}
  \label{fig:dyn_system_hf}
\end{figure}

\begin{figure}[tbp]
  \centering
  \includegraphics[width=5.5in]{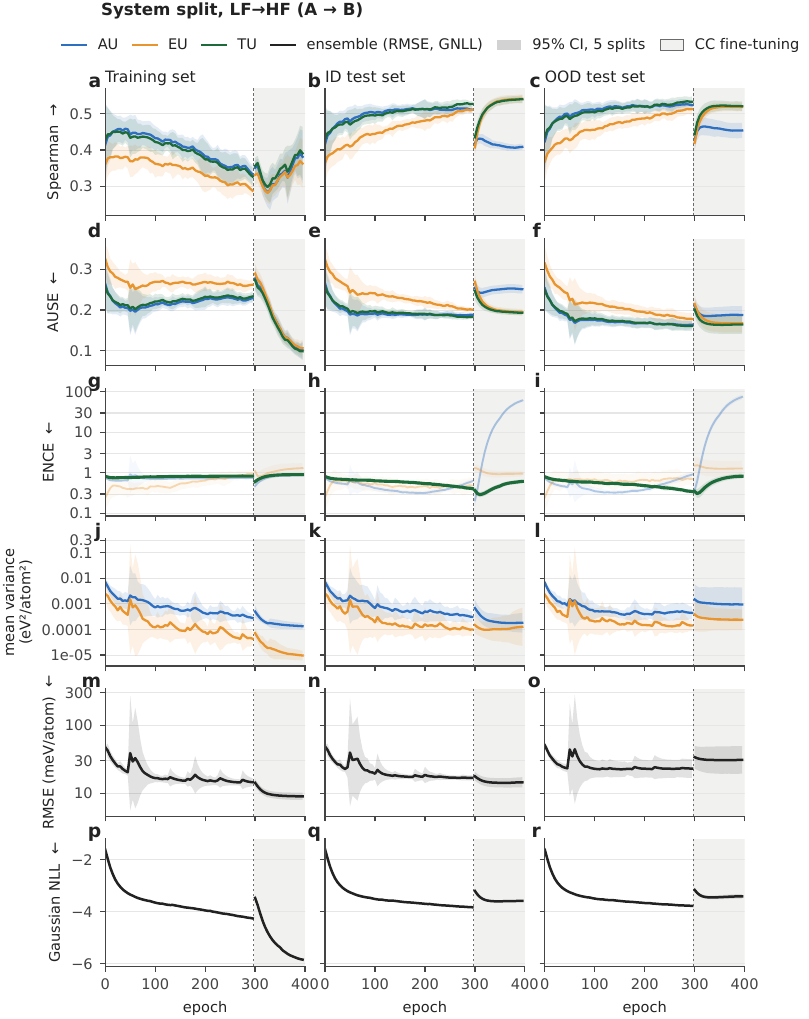}
  \caption{\textbf{Training dynamics of \LFHF\ ensembles under system shift.} As Figure~\ref{fig:dyn_system_hf}. Epochs 0--299 are DFT pretraining (the LF-only ensemble, DFT labels), epochs 300--399 (shaded) CC fine-tuning; RMSE and NLL jump at epoch 300 partly because the target changes.}
  \label{fig:dyn_system_lfhf}
\end{figure}

\begin{figure}[tbp]
  \centering
  \includegraphics[width=5.5in]{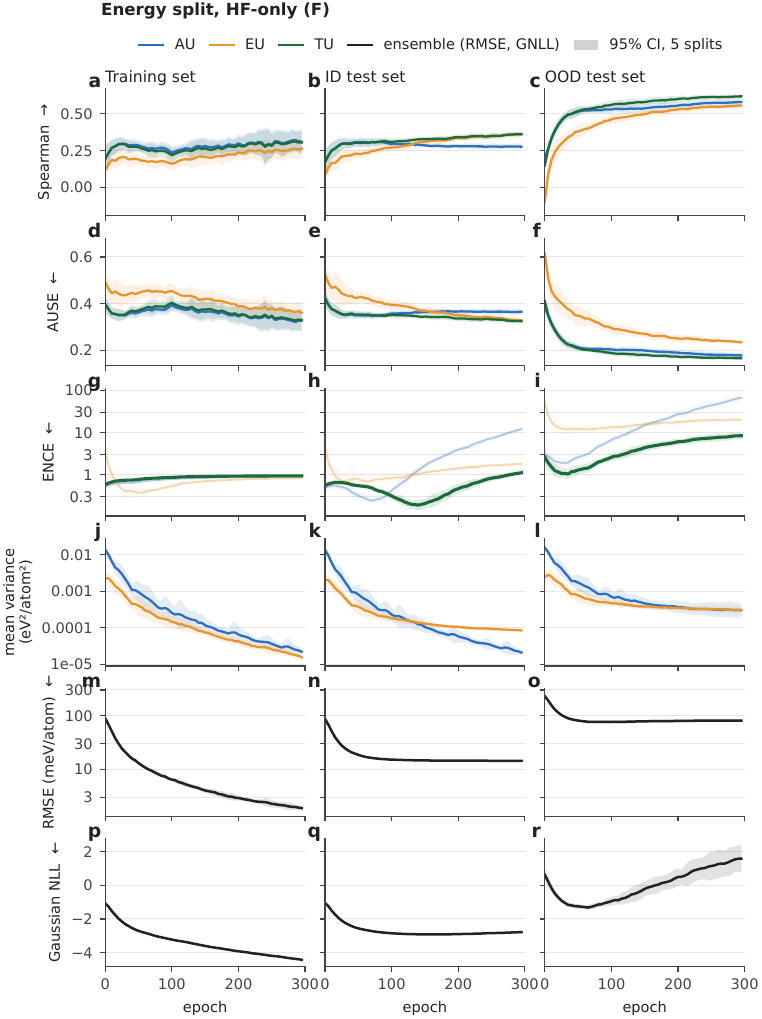}
  \caption{\textbf{Training dynamics of HF-only ensembles under energy shift.} As Figure~\ref{fig:dyn_system_hf}; the Energy-OOD column is support-filtered.}
  \label{fig:dyn_energy_hf}
\end{figure}

\begin{figure}[tbp]
  \centering
  \includegraphics[width=5.5in]{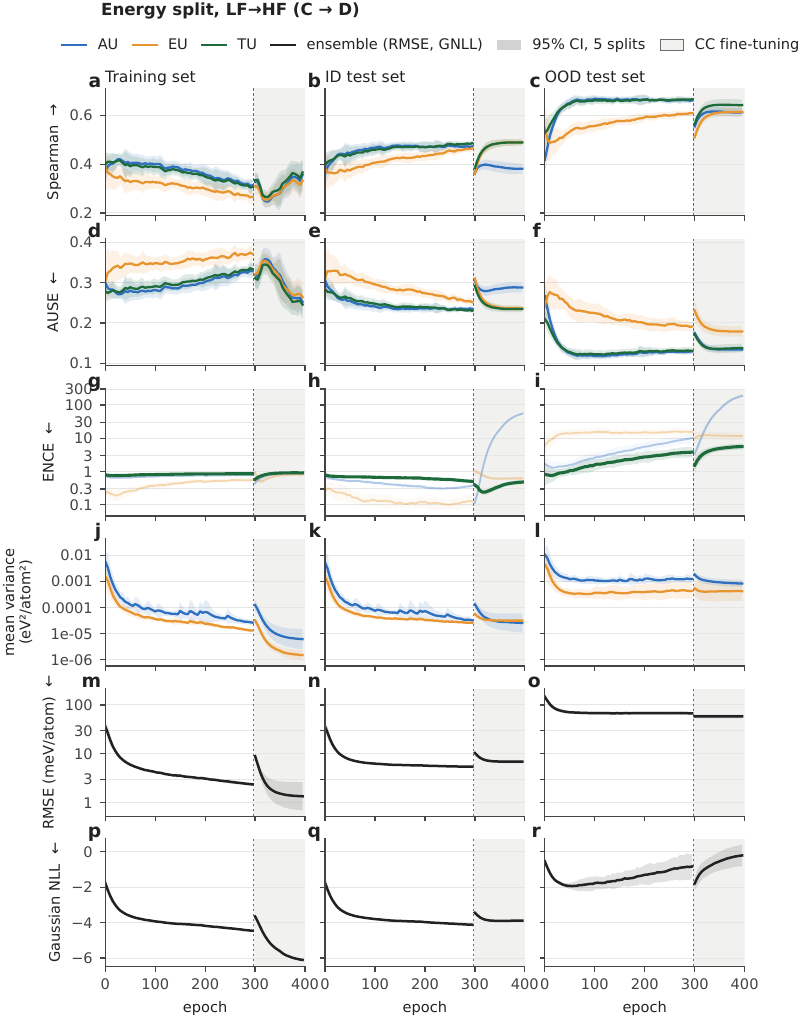}
  \caption{\textbf{Training dynamics of \LFHF\ ensembles under energy shift.} As Figure~\ref{fig:dyn_system_lfhf}; the Energy-OOD column is support-filtered and, during pretraining, scored against DFT labels.}
  \label{fig:dyn_energy_lfhf}
\end{figure}

\begin{figure}[tbp]
  \centering
  \includegraphics[width=5in]{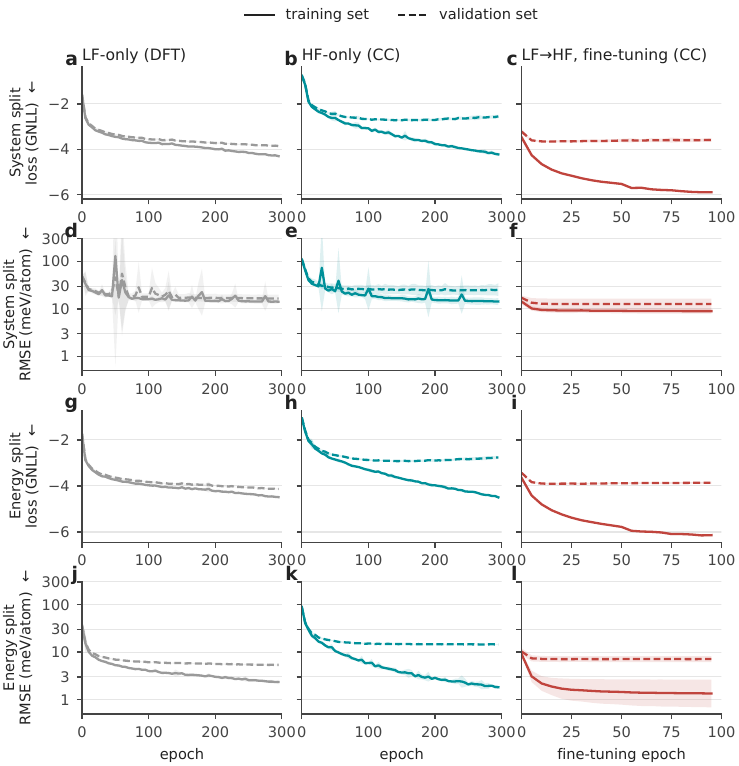}
  \caption{\textbf{Training and validation curves.} Training loss (Gaussian NLL, solid) and validation loss (dashed), and the corresponding RMSE, for LF-only (DFT), HF-only (CC) and the fine-tuning stage of \LFHF\ (CC), under system (top two rows) and energy shift (bottom two rows). Loss: five-split arithmetic means; RMSE: geometric means; bands are 95\% intervals. For HF-only and \LFHF\ the validation loss flattens or rises while the training loss keeps falling, which is why validation-based selection keeps early fine-tuning checkpoints.}
  \label{fig:traincurves}
\end{figure}

This section complements Sections~\ref{sec:results:decomposition} and~\ref{sec:results:hf} with the full per-epoch trajectories of every metric, on the training, ID and OOD sets side by side: HF-only training (Figures~\ref{fig:dyn_system_hf} and~\ref{fig:dyn_energy_hf}) and the whole \LFHF\ trajectory (Figures~\ref{fig:dyn_system_lfhf} and~\ref{fig:dyn_energy_lfhf}), under system and energy shift. The Energy-OOD columns use the training-support filter; the training and ID curves are computed on all configurations, of which the filter would remove at most two per split. Figure~\ref{fig:traincurves} gives the training and validation curves of every experiment.

Four regularities recur. First, ranking on held-out data settles within a few dozen epochs, while calibration keeps changing: during HF-only training the TU calibration error on the ID and OOD sets passes through a minimum and then rises for the rest of the run, driven by the aleatoric component, and during fine-tuning the aleatoric calibration error rises from the first epoch. Second, on the training set the calibration error of every signal stays close to, and below, one throughout, without the rise seen on held-out data. Third, on the training set RMSE and NLL keep decreasing, whereas on the ID and OOD sets they flatten early; the gap opens most clearly during CC fine-tuning, where the training error keeps falling while the held-out error barely moves. This widening gap between training and held-out error is plausibly what the variance head, fitted to the training residuals, fails to capture. Fourth, the validation loss (Figure~\ref{fig:traincurves}) keeps improving over 300 epochs for LF-only training, reaches a minimum after a few dozen epochs for HF-only training and then slowly worsens, and stops improving within the first few epochs of fine-tuning while the training loss keeps falling, which is why validation-based selection keeps such early fine-tuning checkpoints (Appendix~\ref{sec:hyperparameters}). System-split pretraining also shows a transient instability around epochs 50--60 in two of the five splits, visible in every metric and in the EU/AU ratio of Appendix~\ref{sec:supp:ratio}.

\FloatBarrier
\subsection{Balance between epistemic and aleatoric uncertainty}
\label{sec:supp:ratio}

\begin{figure}[h]
  \centering
  \includegraphics[width=\linewidth]{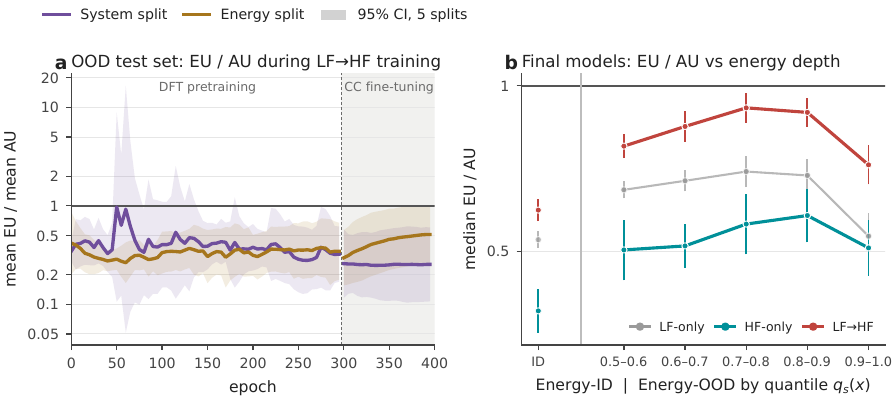}
  \caption{\textbf{EU is smaller than AU at every energy depth and for nearly all of training.} \textbf{(a)} Mean EU over mean AU on the OOD test sets during \LFHF\ training, DFT pretraining (epochs 0--299) then CC fine-tuning (300--399, shaded); bands are 95\% intervals. \textbf{(b)} Converged energy-split models: median per-configuration EU/AU on Energy-ID and in within-system energy-quantile bins of Energy-OOD, for LF-only (grey, DFT labels), HF-only and \LFHF.}
  \label{fig:euau}
\end{figure}

Figure~\ref{fig:euau}a tracks the ratio of mean EU to mean AU on the OOD test sets. Apart from two early pretraining epochs on the system split, where the five-split mean exceeds one with an interval spanning several orders of magnitude, it stays below one. It does not grow during pretraining (energy split $0.42\to0.35$, system split $0.35\to0.33$), and changes only during fine-tuning, rising to $0.51$ on the energy split and settling near $0.26$ on the system split. The wide system-split band during pretraining reflects split-to-split variation in a ratio of two small means. Figure~\ref{fig:euau}b uses the median per-configuration ratio, which is robust to the few configurations with extreme EU. The median EU/AU stays below one in every bin and for every protocol. It rises from Energy-ID to the middle of the tail (\LFHF\ $0.59\to0.91$, HF-only $0.39\to0.58$, LF-only $0.53\to0.70$) and falls again in the highest-energy bin ($0.72$, $0.51$ and $0.53$). \LFHF\ has the largest EU share in every bin, above both single-fidelity baselines.

\section{Supplementary results for the converged models}
\label{sec:supp}

This section complements Section~\ref{sec:results:final}. Statistics follow Section~\ref{sec:experiments}: five-split means with two-sided 95\% Student-t intervals, geometric means with log-space intervals for ENCE and RMSE, and the training-support filter applied throughout.

\subsection{Reliability diagrams}
\label{sec:supp:reliability}

\begin{figure}[tbp]
  \centering
  \includegraphics[width=5.1in]{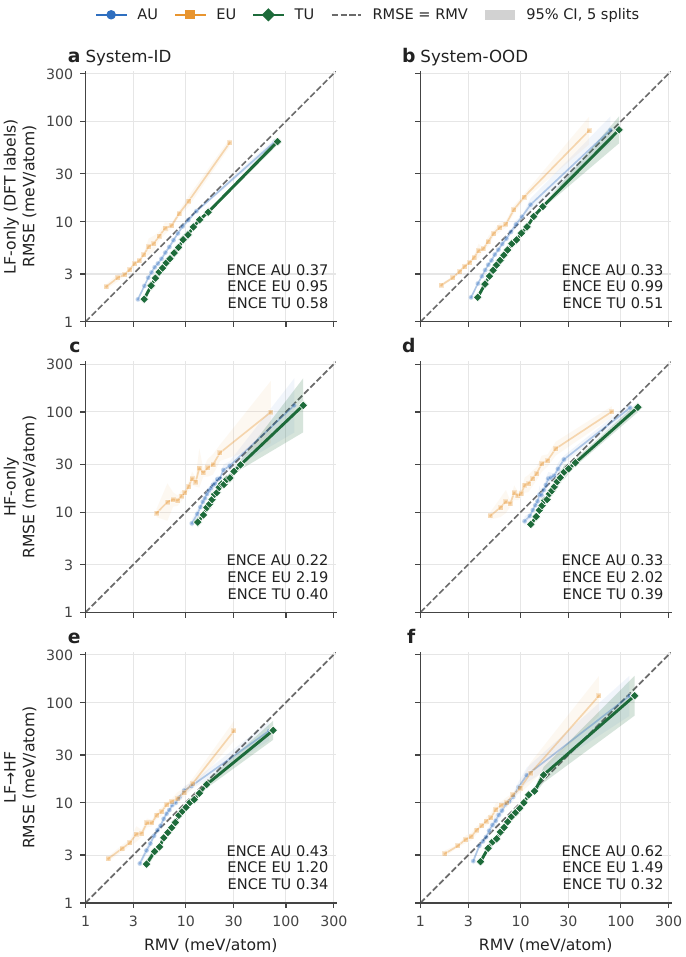}
  \caption{\textbf{Reliability of the converged ensembles under system shift, before recalibration.} Rows: LF-only (DFT labels), HF-only and \LFHF{} (CC labels); columns: ID and OOD test set. In each split, the support-filtered test configurations are sorted by each signal and cut into 15 equal-count bins; each point is the five-split geometric mean of the root mean predicted variance (RMV, $x$) and of the RMSE ($y$) of one bin, per atom, on common log-log axes. Bands are 95\% log-space intervals of the RMSE over five splits (RMV intervals not drawn); the dashed diagonal marks perfect calibration, and points above it are overconfident. The corner text gives the five-split geometric-mean ENCE of each signal.}
  \label{fig:rel_system}
\end{figure}
\begin{figure}[tbp]
  \centering
  \includegraphics[width=5.1in]{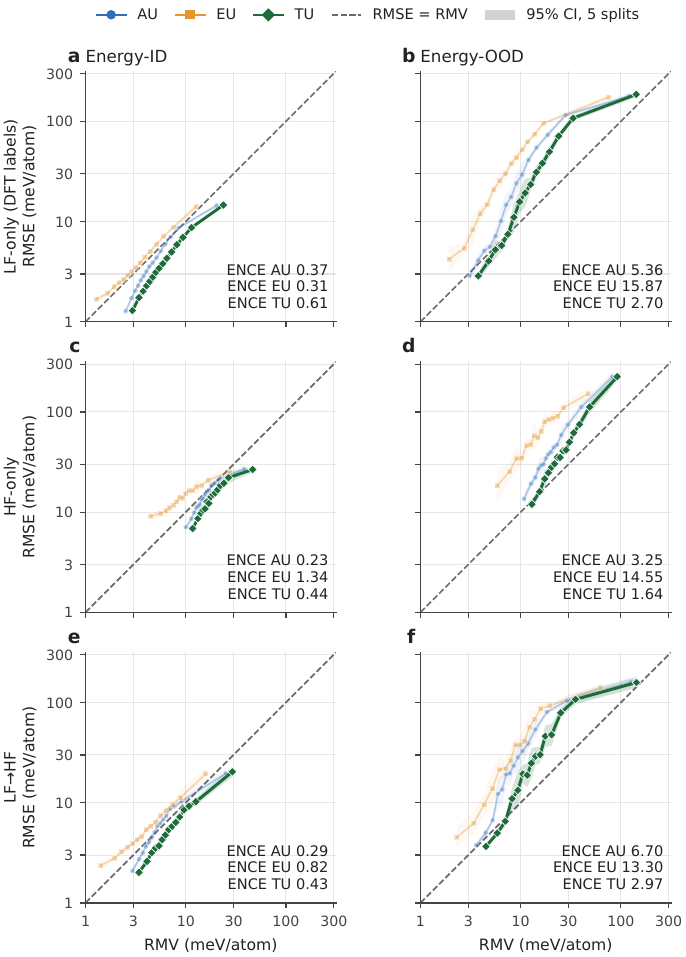}
  \caption{\textbf{Reliability of the converged ensembles under energy shift, before recalibration.} As Figure~\ref{fig:rel_system}.}
  \label{fig:rel_energy}
\end{figure}

Figures~\ref{fig:rel_system} and~\ref{fig:rel_energy} show the reliability diagrams of every converged ensemble, including the LF-only ensembles evaluated against DFT labels, from which the calibration results of Section~\ref{sec:results:final} are computed. On the system split the bands are wide in the few highest-variance bins, which hold a handful of large-error configurations per split; on the energy split they are narrow throughout.

Under system shift (Figure~\ref{fig:rel_system}), the ID and OOD diagrams are nearly indistinguishable for every protocol: TU follows the diagonal or lies slightly below it. Of its components, AU alone behaves similarly, whereas EU alone lies above the diagonal, as expected of one part of the predictive variance. Under energy shift (Figure~\ref{fig:rel_energy}), TU stays close to the diagonal in domain, while on Energy-OOD it lies far above it for every protocol, and so do both of its components. This includes the LF-only ensembles, which suggests that overconfidence under energy extrapolation is not specific to high-fidelity training; high-fidelity training adds the progressive degradation of the aleatoric term (Section~\ref{sec:results:hf}).

\FloatBarrier
\subsection{Rank correlation of the converged models}
\label{sec:supp:spearman}

Table~\ref{tab:spearman} reports the Spearman rank correlation between predicted uncertainty and squared error for the converged models of Figure~\ref{fig:final}. It agrees with AUSE: \LFHF\ has a higher correlation than HF-only for every signal on every test set, EU is the weakest signal without pretraining, and after pretraining EU exceeds AU on System-ID, System-OOD and Energy-ID. Both protocols rank Energy-OOD errors better than Energy-ID errors, as in Figure~\ref{fig:shift}.

\begin{table}[h]
\caption{Spearman rank correlation ($\uparrow$) between uncertainty and squared error of the converged models, five-split mean $\pm$ 95\% half-width.}
\label{tab:spearman}
\begin{center}
\small
\setlength{\tabcolsep}{4pt}
\begin{tabular}{lcccccc}
\toprule
 & \multicolumn{3}{c}{HF-only} & \multicolumn{3}{c}{\LFHF} \\
\cmidrule(lr){2-4}\cmidrule(lr){5-7}
Test set & AU & EU & TU & AU & EU & TU \\
\midrule
System-ID  & $.365\pm.013$ & $.338\pm.010$ & $.377\pm.012$ & $.463\pm.009$ & $.492\pm.013$ & $.500\pm.009$ \\
System-OOD & $.402\pm.045$ & $.370\pm.040$ & $.411\pm.044$ & $.490\pm.025$ & $.501\pm.013$ & $.518\pm.019$ \\
Energy-ID  & $.310\pm.020$ & $.255\pm.020$ & $.315\pm.022$ & $.420\pm.024$ & $.449\pm.015$ & $.457\pm.019$ \\
Energy-OOD & $.533\pm.022$ & $.445\pm.015$ & $.556\pm.018$ & $.626\pm.018$ & $.589\pm.019$ & $.640\pm.019$ \\
\bottomrule
\end{tabular}
\end{center}
\end{table}

\subsection{Post-hoc isotonic recalibration}
\label{sec:supp:recal}

\begin{figure}[h]
  \centering
  \includegraphics[width=0.75\linewidth]{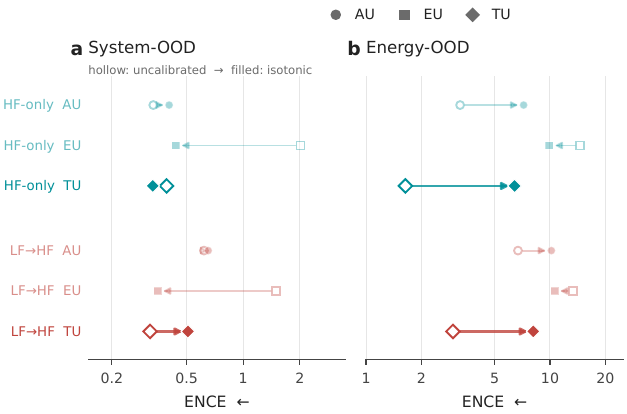}
  \caption{\textbf{Isotonic recalibration fitted in domain transfers poorly to energy extrapolation.} ENCE of the converged models on \textbf{(a)} System-OOD and \textbf{(b)} Energy-OOD before (hollow) and after (filled) isotonic recalibration fitted on the in-domain validation set, per protocol and signal.}
  \label{fig:recal}
\end{figure}

\begin{figure}[tbp]
  \centering
  \includegraphics[width=5.1in]{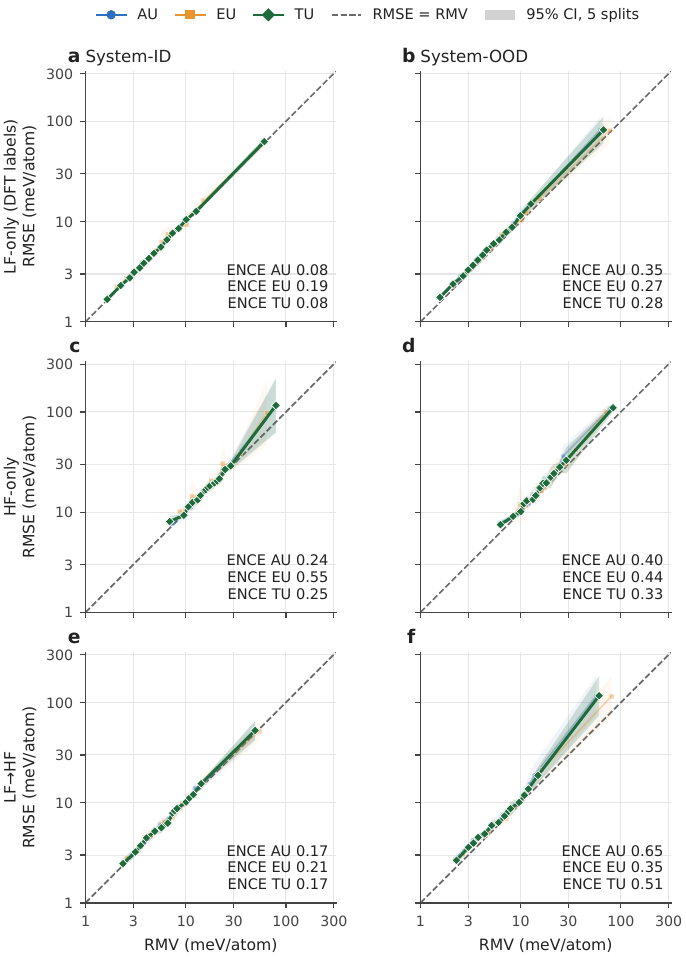}
  \caption{\textbf{Reliability of the converged ensembles under system shift, after isotonic recalibration.} Rows: LF-only (DFT labels), HF-only and \LFHF{} (CC labels); columns: ID and OOD test set. In each split, the support-filtered test configurations are sorted by each signal and cut into 15 equal-count bins; each point is the five-split geometric mean of the root mean predicted variance (RMV, $x$) and of the RMSE ($y$) of one bin, per atom, on common log-log axes. Bands are 95\% log-space intervals of the RMSE over five splits (RMV intervals not drawn); the dashed diagonal marks perfect calibration, and points above it are overconfident. The corner text gives the five-split geometric-mean ENCE of each signal. The isotonic map of each signal is fitted on the in-domain validation set of each split.}
  \label{fig:rel_system_cal}
\end{figure}
\begin{figure}[tbp]
  \centering
  \includegraphics[width=5.1in]{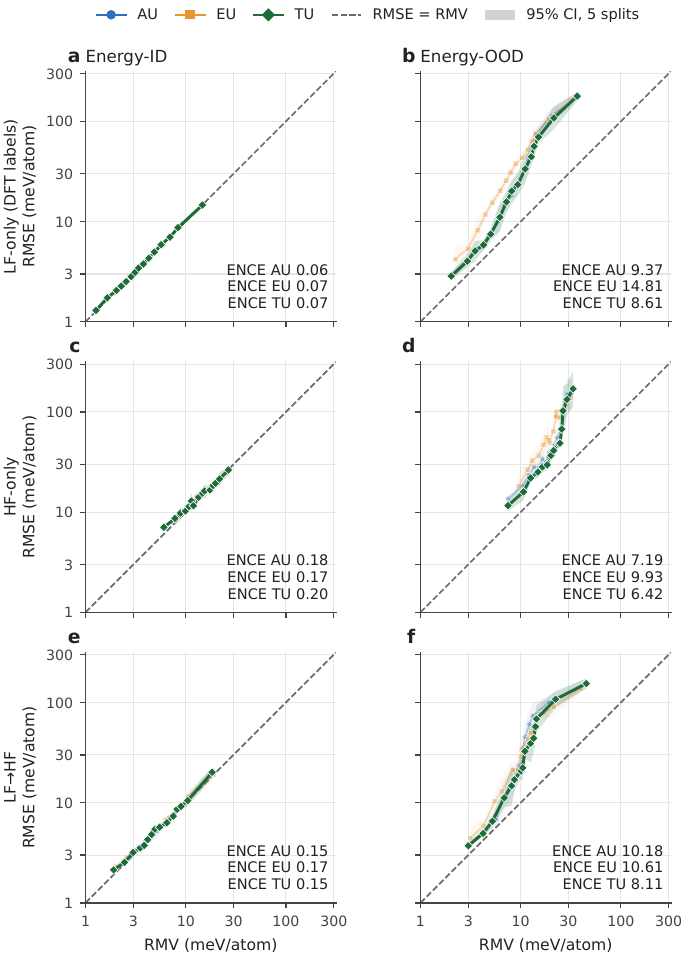}
  \caption{\textbf{Reliability of the converged ensembles under energy shift, after isotonic recalibration.} As Figure~\ref{fig:rel_system_cal}.}
  \label{fig:rel_energy_cal}
\end{figure}

For each signal we fit an isotonic regression from predicted variance to squared error on the ensemble's validation predictions in each split~\citep{kuleshov2018accurate,busk2023graph}, and apply it to the test sets. The map is monotone and is held constant beyond the range of validation variances. Recalibration changes only the predicted variances: the ensemble mean, and hence the RMSE, are those of Figure~\ref{fig:final}, and AUSE and Spearman change only through the ties that the piecewise-constant map introduces. In domain it works as intended, reducing TU ENCE from $0.444$ to $0.196$ (HF-only) and from $0.427$ to $0.150$ (\LFHF) on Energy-ID, and from $0.401$ to $0.250$ and from $0.342$ to $0.174$ on System-ID. Figure~\ref{fig:recal} shows the OOD test sets. Under system shift, recalibration repairs the badly scaled epistemic signal ($2.02\to0.44$ for HF-only, $1.49\to0.35$ for \LFHF), with mixed effects on TU ($0.391\to0.330$ for HF-only, $0.319\to0.508$ for \LFHF). Under energy extrapolation it inflates the calibration error of AU ($3.25\to7.19$ and $6.70\to10.2$) and TU ($1.64\to6.42$ and $2.97\to8.11$) for both protocols, and improves EU only slightly ($14.5\to9.93$ and $13.3\to10.6$). Because the map preserves the ordering up to ties, ranking is essentially unchanged (Energy-OOD TU Spearman $0.556\to0.547$ for HF-only and $0.640\to0.632$ for \LFHF). The failure appears to be one of scale: high-energy errors exceed anything seen in the validation set, and a map that is constant beyond the validation range cannot extend the variance scale to them.

Figures~\ref{fig:rel_system_cal} and~\ref{fig:rel_energy_cal} show the reliability diagrams after recalibration. In domain, all three signals of every protocol collapse onto the diagonal, and on System-OOD they stay close to it. On Energy-OOD the three signals become nearly indistinguishable but remain far above the diagonal, because the recalibrated variances cannot exceed the largest value seen on the validation set.

\FloatBarrier
\section{Acquisition experiment}
\label{sec:app:al}

\paragraph{Protocol.} For each split, the 5{,}000-configuration Energy-OOD CC test set is divided, stratified by molecular system, into an acquisition pool and a held-out test set of 2{,}500 configurations each; the held-out half is not a higher-energy half. The pool labels are hidden. Each strategy selects 500 pool configurations: uniformly at random (one batch per split, shared by both protocols), or the 500 with the largest AU, EU or TU under the converged HF-only or \LFHF\ ensemble. Every member of the corresponding ensemble is then warm-started from its converged checkpoint, with a fresh optimizer and learning-rate schedule, and trained for 50 additional epochs with learning rate $10^{-3}$ on its CC training set plus the 500 acquired configurations. Checkpoints are selected on the original validation set. Gains relative to random selection are computed within each split before averaging over splits. The training-support filter (Appendix~\ref{sec:app:filter}) removes 6--10 held-out configurations per split. In addition, one held-out configuration of one split, on which the \LFHF\ ensemble extrapolated catastrophically, was removed after inspection; that split therefore has 2{,}499 held-out configurations before filtering.

\paragraph{Results.} Figure~\ref{fig:acquisition}a shows the RMSE reduction on the held-out configurations relative to random selection. Uncertainty-guided acquisition beats random selection for all six combinations of protocol and signal. The benefit is larger after pretraining, roughly $2.3$--$2.5$~meV/atom against $0.8$--$1.6$~meV/atom without it, and the difference between the two protocols is resolved for every signal. Without pretraining, EU is the best signal, ahead of AU by $0.80\pm0.55$ and of TU by $0.46\pm0.41$~meV/atom (paired over splits); after pretraining the three signals are indistinguishable. Figure~\ref{fig:acquisition}b shows why uncertainty helps: every signal concentrates its batch on the high-energy end of each molecule's range, where the model is weakest.

\paragraph{In-domain cost.}
\begin{figure}[h]
  \centering
  \includegraphics[width=0.55\linewidth]{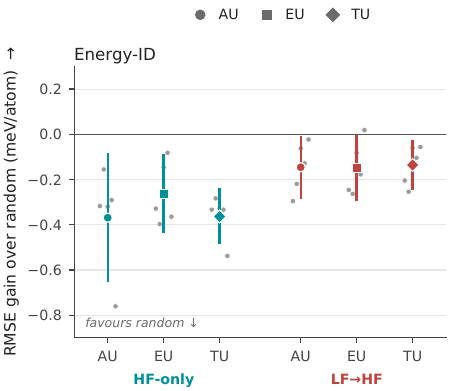}
  \caption{\textbf{Uncertainty-guided acquisition from the Energy-OOD pool costs little in domain.} Energy-ID RMSE reduction relative to random selection of the same budget, for the acquisition experiment of Section~\ref{sec:results:al}; negative values favour random selection. Markers are five-split means with 95\% intervals, grey dots the individual splits.}
  \label{fig:al_id}
\end{figure}

Figure~\ref{fig:al_id} evaluates the acquisition experiment of Section~\ref{sec:results:al} on Energy-ID. All acquired configurations come from the Energy-OOD pool, and continued training improves Energy-ID RMSE for every strategy (random selection: $15.60\to15.06$~meV/atom for HF-only and $7.86\to7.42$~meV/atom for \LFHF). Uncertainty-guided batches improve it slightly less than random ones, consistent with their concentration on the highest-energy configurations (Figure~\ref{fig:acquisition}b), which are furthest from the in-domain region. The cost relative to random selection is $0.37$ [0.08, 0.65] (AU), $0.26$ [0.09, 0.44] (EU) and $0.36$ [0.24, 0.49]~meV/atom (TU) for HF-only, and $0.15$ [0.01, 0.28], $0.15$ [0.00, 0.30] and $0.14$ [0.02, 0.25]~meV/atom for \LFHF. After pretraining the cost is smaller, by $0.23$ [0.02, 0.44]~meV/atom for TU, while for AU and EU the difference between protocols is not resolved with five splits.

\FloatBarrier

\FloatBarrier
\section{Limitations and future work}
\label{sec:limitations}

We focus only on energy prediction. While energies are central to atomistic modeling and provide a clean setting for uncertainty analysis, practical molecular simulations also require accurate and reliable forces. Extending the analysis to force uncertainty is an important direction for future work, especially because force errors are local, vector-valued, and directly affect molecular dynamics stability.

We use mean--variance deep ensembles as our uncertainty estimator. This choice provides a strong, theoretically founded, and widely used baseline for decomposing predictive uncertainty into aleatoric and epistemic components, but it is computationally expensive because it requires training and evaluating multiple models. Future work should investigate whether cheaper alternatives, such as shallow ensembles, last-layer ensembles, multi-head models, evidential methods, or approximate Bayesian approaches, reproduce the same uncertainty dynamics observed here. We also do not ablate the parts of the estimator, for example by training ensembles with a squared loss, using a single mean--variance model, or resetting the variance head at the start of fine-tuning.

Our analysis relies on operational definitions. With deterministic labels, there is no ground truth for the aleatoric and epistemic terms separately, so we judge calibration on TU and use the ENCE of AU and EU only as a diagnostic (Appendix~\ref{sec:app:theory}). The comparison between HF-only and \LFHF\ is one of data access: \LFHF\ also sees the DFT labels, including the geometries of the CC training set, and we do not separate the effect of this additional information from that of the change of fidelity. Likewise, the two shifts use different training sets, so we compare each OOD test set with the ID test set of its own split rather than the two shifts with each other in absolute terms.

Our main experiments are based on ANI-1x and on controlled system-level and energy-based splits. This design is useful for isolating the effect of data fidelity and distributional shift, but it does not capture the full complexity of realistic simulation workflows, where chemical composition, molecular size, temperature, conformation, and energy can shift simultaneously. Evaluating the same questions on larger and more diverse datasets, including materials and biomolecular systems, would test the generality of our conclusions.

\FloatBarrier
\section{Broader impact}
\label{sec:broader_impact}

This work is primarily methodological and focuses on understanding predictive uncertainty in atomistic machine learning models. As such, it does not directly introduce new capabilities for generating data or making decisions affecting individuals or groups, and its societal impact is indirect.

On the positive side, improving the understanding of uncertainty in atomistic models can contribute to more reliable simulations in chemistry and materials science. This has potential downstream benefits in areas such as drug discovery, energy storage, and materials design, where more accurate and trustworthy models can accelerate scientific progress while reducing computational cost. In particular, better uncertainty estimation can enable more efficient data acquisition strategies, reducing the need for expensive quantum-mechanical calculations and lowering the environmental footprint of large-scale simulations.

Potential negative impacts are limited but not entirely absent. More efficient and reliable atomistic modeling tools could be used to accelerate the design of harmful chemical compounds or materials if applied inappropriately. Additionally, incorrect or misinterpreted uncertainty estimates could lead to overconfidence in model predictions, potentially resulting in flawed scientific conclusions or unsafe downstream applications. However, our work explicitly highlights the limitations and regime-dependent behavior of uncertainty estimates, which may help mitigate such risks by discouraging overreliance on poorly calibrated models.

Overall, we believe the primary impact of this work is to improve the transparency and reliability of uncertainty estimation in atomistic machine learning, contributing to safer and more informed use of these models in scientific and engineering applications.

\end{document}